\documentclass{article}

\usepackage{microtype}
\usepackage{graphicx}
\usepackage{subfigure}
\usepackage{booktabs} % for professional tables
\usepackage{multirow}
\usepackage{xcolor}
\usepackage{float}
\definecolor{textgray}{gray}{0.6} % 定义一种柔和的灰色% If your build breaks (sometimes temporarily if a hyperlink spans a page)
\usepackage{hyperref}

\usepackage[section]{placeins}  % 提供 \FloatBarrier

\usepackage[preprint]{icml2026}

\usepackage{amsmath}
\usepackage{amssymb}
\usepackage{mathtools}
\usepackage{amsthm}

\usepackage[capitalize,noabbrev]{cleveref}

\theoremstyle{plain}
\newtheorem{theorem}{Theorem}[section]
\newtheorem{proposition}[theorem]{Proposition}

\theoremstyle{definition}

\theoremstyle{remark}

\usepackage[textsize=tiny]{todonotes}

\icmltitlerunning{Multi-scale Graph Autoregressive Modeling: Molecular Property Prediction via Next Token Prediction}

\begin{document}

\twocolumn[
    \icmltitle{Multi-scale Graph Autoregressive Modeling: Molecular Property Prediction via Next Token Prediction}
    
    \begin{icmlauthorlist}
    \icmlauthor{Zhuoyang Jiang}{hkust(gz),zgca}
    \icmlauthor{Yaosen Min}{zgca}
    \icmlauthor{Peiran Jin}{zgca}
    \icmlauthor{Lei Chen}{hkust(gz)}
    \end{icmlauthorlist}
    
    \icmlaffiliation{hkust(gz)}{The Hong Kong University of Science and Technology (Guangzhou), Guangzhou, China}
    \icmlaffiliation{zgca}{Zhongguancun Academy, Beijing, China}
    
    \icmlcorrespondingauthor{Yaosen Min}{yaosenmin@zgci.ac.cn}
    \icmlcorrespondingauthor{Peiran Jin}{jinpeiran@bjzgca.edu.cn}
    \icmlkeywords{Graph Representation Learning, Autoregressive Modeling, Self-Supervised Learning, Science Foundation Models, Molecular Property Prediction}
    \vskip 0.3in
]

\printAffiliationsAndNotice{} % otherwise use the standard text.

\begin{abstract}
We present \textbf{Connection-Aware Motif Sequencing (CamS)}, a graph-to-sequence representation that enables decoder-only Transformers to learn molecular graphs via standard next-token prediction (NTP).  For molecular property prediction, SMILES-based NTP scales well but lacks explicit topology, whereas graph-native masked modeling captures connectivity but risks disrupting the pivotal chemical details (e.g., activity cliffs). CamS bridges this gap by serializing molecular graphs into structure-rich causal sequences. CamS first mines data-driven connection-aware motifs. It then serializes motifs via scaffold-rooted breadth-first search (BFS) to establish a stable core-to-periphery order. Crucially, CamS enables hierarchical modeling by concatenating sequences from fine to coarse motif scales, allowing the model to condition global scaffolds on dense, uncorrupted local structural evidence. We instantiate \textbf{CamS-LLaMA} by pre-training a vanilla LLaMA backbone on CamS sequences. It achieves state-of-the-art performance on MoleculeNet and the activity-cliff benchmark MoleculeACE, outperforming both SMILES-based language models and strong graph baselines. Interpretability analysis confirms that our multi-scale causal serialization effectively drives attention toward cliff-determining differences.
% The code is available at \url{https://anonymous.4open.science/r/CamS-F3B9/}.
\end{abstract}

\section{Introduction}
% [Part 1: Standardization & Data-Centric Motivation]
Molecular property prediction is a core challenge in drug discovery~\cite{wu2018moleculenet} and has increasingly become a primary benchmark for ``Foundation Models (FMs) of science.''~\cite{khan2025comprehensive}
Within FM paradigm, the training recipe has standardized: \textbf{decoder-only Transformers trained via next-token prediction (NTP)} serve as the dominant engine for scale and generalization~\cite{wang2024emu3, xia2025nature, zhang2025unigenx}. 
Consequently, as the core components of these training stacks become heavily optimized, the pragmatic research focus shifts away from bespoke backbone re-design toward optimizing the \textbf{input representation} to fit this proven architecture~\cite{touvron2023llama}. 
However, a critical gap remains: current NTP-based molecular models, which largely rely on 1D SMILES strings, fail to explicitly capture graph topology and consequently trail behind specialist graph models in predictive accuracy~\cite{xia2025nature}.

% [Part 2: The Logic of "Fixing" vs. "Switching"]
Conversely, graph-native methods explicitly model topology but typically \textbf{depart from pure decoder-only NTP}, instead adopting hybrid architectures or objectives that involve, to varying degrees, \textbf{input corruption} (e.g., masked node prediction)~\cite{li2023knowledge, shehzad2024graph, lu2025uni, kong2025unimomo}. 
While sophisticated strategies---such as incorporating global knowledge nodes or avoiding random masking---have been developed to \textbf{mitigate} the information loss caused by masking~\cite{li2023knowledge, rong2020self, liu2024mask, you2020graph, xu2021self}, these methods remain fundamentally bound by the corruption-coverage trade-off \cite{wettig2023should}. 
This limitation is particularly acute in \textit{activity-cliff} scenarios, where a single atomic change triggers a drastic property shift; masking the local neighborhood of a critical functional group effectively removes the precise evidence needed to discern such subtle differences.
Standard NTP, by contrast, offers a dense supervision signal without corrupting the visible prefix context~\cite{clark2020electra}, yet it lacks a \textbf{representation interface} that exposes graph connectivity as effectively as text.

To bridge this gap, we propose \textbf{Connection-Aware Motif Sequencing (CamS)}, a tokenizer-level interface that makes molecular graphs directly learnable by standard decoder-only NTP through a three-stage design.
First, to ensure information efficiency while preserving semantics, we adapt byte-pair encoding (BPE)-style mining on molecular graphs~\cite{shibata1999byte, geng2023novo, shen2024graphbpe}. This produces motif vocabulary with frequent merge operation and introduces a tunable molecular scale that allows direct granularity control. To ensure faithful encoding of fine-grained chemical detail, we further integrate a Single-Atom Vocabulary Closure (SAVC) mechanism.
Second, to make the graph autoregressive-ready, we partition the molecule into a BPEGraph at a specific scale and serialize it via scaffold-rooted breadth-first search (BFS) ~\cite{bundy1984breadth}. This produces a CamS subsequence with a stable \textbf{Intra-scale Order}, moving from the global core to peripheral functional groups.
Third, and most critically, to construct a hierarchical context, we concatenate subsequences from fine to coarse. This strategy circumvents the trade-off between scales and establishes an \textbf{Inter-scale Order} in the final CamS sequence, enabling high-level motifs to be predicted by conditioning on dense, uncorrupted fine-scale evidence.
Collectively, these steps unify substructure compression and serialization through a \textbf{dual-causal ordering}—sequencing motifs within each scale and stacking scales hierarchically—to enable the model to comprehend global scaffolds based on explicit, uncorrupted local structural evidence.

Built on CamS, we instantiate \textbf{CamS-LLaMA} by pre-training a native decoder-only backbone with standard NTP on CamS sequences, keeping the architecture and objective unchanged \cite{touvron2023llama}. 
To keep attribution controlled, we inject only a lightweight molecular fingerprint \cite{cereto2015molecular} prior during downstream fine-tuning, while keeping pre-training purely sequence/structure-driven. 
Across MoleculeNet \cite{wu2018moleculenet} and the activity-cliff stress test MoleculeACE \cite{van2022exposing}, CamS-LLaMA achieves state-of-the-art (SOTA) performance at a comparable model scale to strong graph self-supervised learning baselines, despite relying on substantially weaker priors \cite{li2023knowledge}. 
Beyond accuracy, we provide mechanism-level evidence. Targeted \textbf{ablation studies} confirm that multi-scale concatenation is the key driver of performance, while \textbf{interpretability} analysis reveals that the causal serialization effectively drives attention toward cliff-determining structural differences.

\noindent\textbf{Contributions.}
(1) \textbf{Methodological Framework}: We introduce CamS, a unified graph-to-causal-sequence interface that resolves the conflict between graph topology preservation and scalable autoregressive (AR) training, enabling NTP to function as a structure-native objective.
(2) \textbf{Empirical Validation}: We demonstrate that CamS-LLaMA achieves SOTA performance on MoleculeNet and MoleculeACE benchmarks. Mechanism analysis reveals that the multi-scale causal context explicitly improves attention focus on subtle, activity-determining structural edits.
(3) \textbf{Implementation Recipe}: We provide a reproducible pipeline—from the CamS-tokenizer to the CamS-LLaMA—establishing a robust baseline for applying standard AR FM architectures to molecular science.

\section{Related Work}
\subsection{Molecular Property Prediction via FMs}
Molecular property prediction is increasingly adopting a FM approach: large-scale Transformer pre-training followed by fine-tuning for diverse property endpoints \cite{awais2025foundation, xia2025nature}. 
One approach adopts standard large language model (LLM) methodology by representing molecules as SMILES strings and pre-training decoder-only models via NTP \cite{xia2025nature, cai2025chemfm}. This leverages mature LLM infrastructure and enables straightforward large-scale training. However, SMILES is not structure-native, weakening topology-based signals. String-based edits often misalign with actual chemical changes, and performance consistently trails strong graph-native methods on property prediction tasks \cite{xia2025nature}. 
Alternatively, graph-native FMs directly process molecular graphs to preserve connectivity and local topology \cite{you2021graph, liupre, you2020graph, xu2021self}. High-performing methods typically employ encoder-only Graph Transformers \cite{shehzad2024graph} designed for property prediction. While effective, these models deviate from the vanilla decoder-only NTP approach that exhibits superior generalization and scalability \cite{li2023knowledge, rong2020self}. 
More recent structural FMs integrate graphs with AR components but frequently depend on additional objectives (often incorporating 3D structure) and learned molecular encoders rather than pure tokenizer-level NTP \cite{kong2025unimomo, lu2025uni, zhang2025unigenx}.

\subsection{Molecular Tokenization Strategies}
Within a FM paradigm, tokenizer design becomes a key interface for scaling backbones. String-based schemes operate on SMILES \cite{weininger1988smiles, wang2019smiles} or SELFIES \cite{krenn2022selfies}.
Graph-based approaches employ diverse substructure definitions: rule-based tokenizations rely on limited handcrafted chemistry rules \cite{zhang2021motif}; 
triplet tokenizations, as used in \textbf{KPGT (Knowledge-guided Pre-training of Graph Transformer)}, show that enriching a node token with only one extra atom can substantially boost property prediction \cite{li2023knowledge}; 
ring-or-path-based \cite{wollschlager2024expressivity} and tree-based \cite{jin2018junction} schemes also expose substructures at different granularities; and other works learn tokenization via trainable rules \cite{sun2024representing} or graph neural networks \cite{liu2023rethinking, luonode}. 
Particularly noteworthy are data-driven approaches that adapt BPE idea to graphs to obtain reusable fragments with strong compression and minimal reliance on rules or additional trained models \cite{geng2023novo, shen2024graphbpe, kong2022molecule}. In addition, cross-scale schemes, which combine motifs at different structural levels, also provide complementary design ideas \cite{ji2022relmole, chenhierarchical}.

\section{Method}
\label{sec:method}
To align chemical fidelity with AR compatibility, we propose \textbf{Connection-Aware Motif Sequencing (CamS)}. This interface serializes molecular graphs into multi-scale, causal sequences naturally consumable by decoder-only models (Section~\ref{sec:method_tokenizer}). 
Instantiated on this representation, \textbf{CamS-LLaMA} performs standard NTP pre-training, injecting a lightweight fingerprint prior only during downstream fine-tuning (Section~\ref{sec:method_cams_llama}). 
We conclude by theoretically contrasting this CamS-LLaMA framework with Graph Transformers paradigm (Section~\ref{sec:method_analysis}).

\subsection{CamS-Tokenizer}
\label{sec:method_tokenizer}
The CamS-tokenizer framework bridges graph topology and scalable NTP through data-driven fragment compression at the substructure level and hierarchical serialization (both intra- and inter-scale) at the molecule level. Full implementation details are provided in Appendix~\ref{app:tokenizer_details}.

\noindent\textbf{CamS Vocabulary Construction.} 
On the substructure level, we adopt a data-driven strategy grounded in the philosophy that compression implies understanding.
% BPE-style Mining
Unlike rigid manual rules, we apply \textbf{BPE-style Mining} to capture statistically significant chemical semantics. Following \cite{geng2023novo, shen2024graphbpe}, we first learn an ordered list of merge operations $\mathcal{O}$, sorted by descending frequency based on co-occurrence statistics.
Since BPE merges are iterative, later operations in $\mathcal{O}$ (lower frequency) build upon earlier ones to combine smaller fragments into larger complex structures. Thus, the list order represents a trajectory from fine-grained local substructures to coarse-grained global scaffolds.
Applying the full list $\mathcal{O}$ yields the BPEMotif set (frequent subgraphs vocabulary) and the base BPE-SAV (single atoms vocabulary) naturally captured during mining.

% Single-Atom Vocabulary Closure (SAVC)
Crucially, distinct from prior works, we introduce \textbf{Single-Atom Vocabulary Closure (SAVC)} to address a critical flaw: while generation operates within a closed vocabulary, predictive encoding encounters unseen atom states. Standard tokenizers greedily back off to \texttt{[UNK]} when specific atom-connectivity forms are absent, erasing element details and creating ``Causal Information Breakpoints'' in the AR stream. To prevent this, SAVC constructs the BPE-SAV Back-off by: (1) enumerating connection-aware tokens for all typical valences, and (2) mapping rare atypical forms of element X to \texttt{[X\_AltForm]} rather than \texttt{[UNK]}. 
% → CamS Vocabulary
The final \textbf{CamS Vocabulary $\Sigma$} is the rigorous union of BPEMotif, the complete SAV (BPE-SAV $\cup$ Back-off), and special tokens. This design not only guarantees that every atom is covered by either a high-level motif or a precise single-atom descriptor but also provides a pivotal manipulable dimension for downstream representation: the \textbf{Motif Scale $s$}.
Formally, we define the Motif Scale $s$ as the \textbf{effective vocabulary size} induced by a specific granularity level. Each scale $s$ corresponds to a specific prefix length $k$ of the merge list $\mathcal{O}$. A small $s$ (requiring fewer merges) activates only high-frequency motifs, keeping the graph fragmented (Fine Scale); conversely, a large $s$ (using more merges) includes rare, late-stage operations that aggregate these fragments into larger scaffolds (Coarse Scale).

\noindent\textbf{Per-Scale Encoding.}
On the molecule level, utilizing the constructed CamS Vocabulary $\Sigma$, we can transform raw atom nodes into discrete substructure nodes containing rich connectivity, contextual compression and controllable scale.
% 1. construct the BPEGraph
Specifically, given a molecular graph $G_{\text{mol}}$ and a target Motif Scale $s$ (associated with the operation prefix $\mathcal{O}_{\le k}$), we apply the corresponding $k$ merge operations to $G_{\text{mol}}$. This process induces a partition of atoms into non-overlapping fragments, forming the BPEGraph $G_s=(V_s, E_s)$. Here, nodes $V_s \subset \Sigma$ denote connection-aware motifs/atoms (ranging from fine fragments to coarse scaffolds depending on $s$) and $E_s$ represents the chemical bonds preserving connectivity between them.
Subsequently, a key challenge remains in how the NTP perceives this high-quality BPEGraph: the definition of causal order. While graphs permit various traversals (e.g., random walks~\cite{lovasz1993random}), we establish a deterministic \textbf{Intra-scale Order}. %(visual examples in Appendix~\ref{app:bpegraph_vis}) 
% 2. Scaffold-Rooted BFS
We select the motif with the largest atom count (typically the core scaffold) as the root and serialize $V_s$ into an ordered list via \textbf{Scaffold-Rooted BFS}. This strategy prioritizes global backbone structure before local substitutions \cite{zhang2021motif}, establishing a stable Center-to-Periphery causal order.
% 3. ID Extraction
Finally, we apply ID Extraction to map ordered nodes to their indices in $\Sigma$, yielding the single-scale \textbf{CamS subsequence} $X^{(s)} = (x^{(s)}_1, \dots, x^{(s)}_{L_s})$.

\noindent\textbf{Cross-Scale Concatenation.}
The manipulable Motif Scale $s$ introduces an inherent resolution trade-off: coarse scales capture global scaffolds but offer low resolution, while fine scales preserve atomic details but fragment high-level pharmacophores. 
CamS resolves this by constructing a multi-scale causal context. We tokenize the molecule at a sequence of $M$ increasing scales $\mathcal{S} = \{s_1, s_2, \dots, s_M\}$. 
The resulting CamS Subsequences $\{X^{(s)} \mid s \in \mathcal{S}\}$ are concatenated in a fine-to-coarse order (i.e., the \textbf{Inter-Scale Order}) to form the final \textbf{CamS sequence} $\mathbf{X}$:

\begin{equation}
\label{eq:multi_scale_sequence}
\begin{aligned}
  \mathbf{X} =[ &\,\texttt{[BOS]}, X^{(s_1)}, \texttt{[CONCAT]}, X^{(s_2)}, \dots, \\
   &\,\texttt{[CONCAT]}, X^{(s_M)}, \texttt{[EOS]}].
\end{aligned}
\end{equation}

During training, this \textbf{CamS token stream} allows the model to leverage high-resolution local topology (fine scales) as a prefix to condition the prediction of global scaffolds (coarse scales). This effectively embeds bottom-up structural composition into the AR objective without architectural modifications (analysis in Section~\ref{sec:method_analysis}).

\begin{figure*}[ht]
\vskip 0.1in
\begin{center}
\centerline{\includegraphics[width=\textwidth]{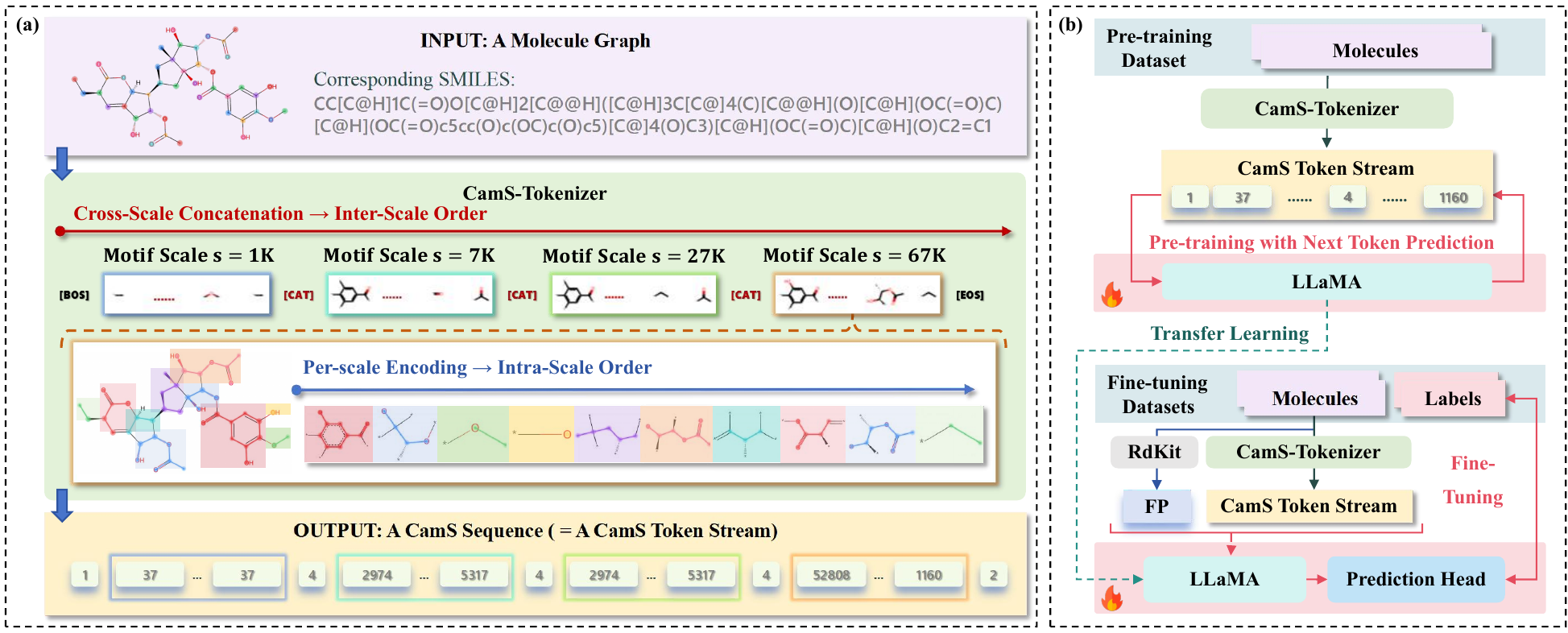}} 
\caption{\textbf{The overall framework of CamS.} 
\textbf{(a) CamS-Tokenizer:} Transforming molecules into causal sequences via the manipulable Motif Scale $s$. 
First, Per-scale Encoding applies merge operations to construct motif graphs, which are serialized via Scaffold-rooted BFS to establish Intra-Scale Order. 
Subsequently, Cross-Scale Concatenation arranges these views from fine ($s=1\mathrm{K}$) to coarse ($s=67\mathrm{K}$) to establish Inter-Scale Order. The resultant CamS sequence serves as a native token stream for a vanilla LLaMA backbone.
\textbf{(b) CamS-LLaMA:} The model is pre-trained on the resultant CamS Token Stream via standard NTP. Pre-trained weights are transferred, and a lightweight fingerprint prior is injected during fine-tuning.
}
\label{fig:framework}
\end{center}
\vskip -0.1in
\end{figure*}

\subsection{CamS-LLaMA}
\label{sec:method_cams_llama}
Following a representation-model co-design principle, CamS-LLaMA leverages the graph-to-sequence interface established in Section~\ref{sec:method_tokenizer} to enable standard AR modeling on molecular graphs. 
As illustrated in Figure~\ref{fig:framework}(a), the multi-scale CamS sequence acts as a transparent bridge, exposing topology to a vanilla LLaMA backbone without architectural modifications. 
This facilitates a scalable two-stage FM pipeline (Figure~\ref{fig:framework}(b)): (1) self-supervised pre-training via NTP to capture structural generative laws, and (2) property-guided fine-tuning with a controlled fingerprint prior.

\noindent\textbf{Autoregressive Pre-training via NTP.} 
CamS-LLaMA adopts a standard LLaMA-style decoder \cite{touvron2023llama}. In pre-training, CamS token lists are treated as ordinary sequences in the shared vocabulary $\Sigma$ and consumed under the usual causal mask.
Concretely, to maximize structural learning signals via data augmentation, each molecule yields five sequence views (four single-scale + one multi-scale); we treat each view as an independent NTP training instance (details in Appendix~\ref{app:multiscale_concat}).
Conditioned on the prefix context reshaped by fine-to-coarse concatenation (Eq.~\eqref{eq:multi_scale_sequence}), the model learns to predict the next token $x^\star_{t+1}$.
The NTP loss is minimized on the pre-training corpus $\mathcal{D}_{\mathrm{pre}}$:
\begin{equation}
  \mathcal{L}_{\mathrm{NTP}}(\theta) = -\mathbb{E}_{X\sim\mathcal{D}_{\mathrm{pre}}} \sum_{t\in\mathcal{I}_{\mathrm{tok}}}\log p_\theta(x^\star_{t+1}\mid x_{\le t}).
  \label{eq:ntp_loss}
\end{equation}
Crucially, because coarse-scale tokens appear after fine-scale ones in $X$, prediction errors propagate gradients through the entire fine-scale prefix, effectively providing a \textbf{fine-to-coarse causal credit assignment} that embeds bottom-up compositional logic into the standard decoder.

\noindent\textbf{Fine-tuning with Fingerprint Injection.}
Many methods inject rich handcrafted descriptors early during pre-training~\cite{li2023knowledge}. In contrast, we keep CamS as the primary representation and inject a lightweight fingerprint prior only during fine-tuning via a dual-path strategy (Early Injection + Late Fusion).
Given a molecule's fingerprint $\mathbf{f} \in \mathbb{R}^{D_{\mathrm{fp}}}$, we project it to $\mathbf{f}' = W_{\mathrm{fp}}\mathbf{f}$ using a learnable projection $W_{\mathrm{fp}} \in \mathbb{R}^{H \times D_{\mathrm{fp}}}$.
First, via Early Injection, $\mathbf{f}'$ is prepended to the input embeddings as a global prompt:
\begin{equation}
  \mathbf{H}^{(0)}_{\mathrm{ft}} = [\,\mathbf{f}'; \mathbf{e}_{\text{BOS}}; \dots; \mathbf{e}_{\text{EOS}}\,] \in \mathbb{R}^{(T+2)\times H}.
  \label{eq:ft_input}
\end{equation}
After $L$ layers, we extract the representation of the \texttt{[EOS]} token, $\mathbf{h}_{\text{EOS}}^{(L)}$, which aggregates the AR structural context.
Second, to allow the backbone to focus on deep implicit reasoning (by offloading explicit feature extraction to the direct path), we perform Late Fusion by concatenating $\mathbf{h}_{\text{EOS}}^{(L)}$ with the original projected fingerprint $\mathbf{f}'$ to form the final representation $\mathbf{u}$:
\begin{equation}
  \mathbf{u} = [\,\mathbf{h}_{\text{EOS}}^{(L)} \,;\, \mathbf{f}'\,] \in \mathbb{R}^{2H}.
  \label{eq:late_fusion}
\end{equation}
This shortcut not only preserves chemical fidelity but also serves as a \textbf{stabilizer} for fine-grained decision making (as analyzed in Section~\ref{sec:ablation}), ensuring predictions remain grounded even when deep features fluctuate. 
This fused vector is fed into task-specific heads. For classification ($\hat{\mathbf{y}}$) and regression ($\hat{y}$), the objectives are:
\begin{equation}
\label{eq:ft_loss}
\begin{aligned}
\mathcal{L}_{\mathrm{cls}} &= - \frac{1}{B} \sum_{i=1}^{B}\log \big(\mathrm{softmax}(W_2 \,\sigma(W_1 \mathbf{u}))\big)_{y^{(i)}}, \\
\mathcal{L}_{\mathrm{reg}} &= \frac{1}{B} \sum_{i=1}^{B}\big(W_2 \,\sigma(W_1 \mathbf{u}) - y^{(i)}\big)^2.
\end{aligned}
\end{equation}
This strategy balances the pre-trained structural knowledge with the explicit fingerprint prior, optimizing $\mathbb{E}_{\mathcal{D}_{\mathrm{task}}}[\mathcal{L}_{\mathrm{task}}]$.

\subsection{CamS LLaMA vs.\ Graph Transformer}
\label{sec:method_analysis}

We analyze CamS-LLaMA and Graph Transformer-style models (e.g., KPGT) within a unified framework of Token-Level Graph-Structured Deep Feature Construction. We show that CamS yields (1) a denser multi-view direct supervision signal and (2) a hierarchical inductive bias that are mathematically distinct from standard masked-node prediction (MNP). Detailed derivations are in Appendix~\ref{app:theory}.

\noindent\textbf{Unified Framework.}
Let $G=(V,E)$ be a molecular graph. Both paradigms construct a token graph $\mathcal{G}_{\mathrm{tok}}=(\mathcal{V}_{\mathrm{tok}},\mathcal{E}_{\mathrm{tok}})$ and perform $L$ layers of propagation:
\begin{equation}
\label{eq:unified_prop}
\begin{split}
    \mathbf{H}^{(l+1)} &= \mathrm{Agg}\big( \mathbf{A}^{(l)} \,\mathrm{Val}(\mathbf{H}^{(l)}) \big), \\
    \mathbf{A}^{(l)}_{ij} &\propto \exp\left(\frac{\mathbf{q}_i^\top \mathbf{k}_j}{\sqrt{d}} + \psi_{ij}\right).
\end{split}
\end{equation}
where $\mathbf{H}^{(l)}$ are token embeddings and $\psi_{ij}$ encodes structural bias.
For Graph Transformers (Flat/Static Bias), $\mathcal{V}_{\mathrm{tok}}$ corresponds to single-scale atoms or triplets. $\psi_{ij}$ encodes \textbf{static} biases (e.g., shortest-path distance) independent of token content, yielding an isotropic receptive field~\cite{ying2021transformers}.
For CamS-LLaMA (Hierarchical/Causal Flow), $\mathcal{V}_{\mathrm{tok}}$ comprises multi-scale connection-aware motifs $\{v^{(s)}\}$. $\psi_{ij}$ is determined by the causal mask $\mathbf{M}_{\mathrm{causal}}$, enforcing a directed flow where coarse-scale tokens attend to fine-scale predecessors (via Inter-Scale Order). This turns cross-scale aggregation into a learned, anisotropic connectivity.

\noindent\textbf{Information-Theoretic Context Analysis.}
We contrast NTP with MNP. Let $x_t$ be a target token and $Z_t$ its uncorrupted evidence set. MNP observes a stochastically corrupted context $\tilde{Z}_t=\mathcal{M}(Z_t)$.
\begin{proposition}[Context Information Inequality]
\label{prop:info_gain}
For any non-trivial masking channel $\mathcal{M}$, we have $I(x_t; Z_t) \geq I(x_t; \tilde{Z}_t)$.
\end{proposition}
\vspace{-0.5em}
\noindent\textit{Sketch.} The inequality follows from the Data Processing Inequality (DPI): masking is a stochastic post-processing of $Z_t$ that cannot increase mutual information~\cite{cover1999elements}. In graph MNP, this loss of information is exacerbated by evidence-pattern uncertainty: the most relevant local-neighborhood evidence for a masked token may be co-masked, making the conditional distribution unstable across masking patterns. Recent graph pretraining works explicitly mitigate this by masking structured units (e.g., triplets or local subgraphs), designing structure-guided masking, or injecting global knowledge tokens~\cite{li2023knowledge, rong2020self, liu2024mask}. In contrast, CamS NTP for coarse-scale tokens conditions on an uncorrupted fine-to-coarse history, avoiding such stochastic evidence loss. This perspective is consistent with classic critiques of Masked Language Modelling (MLM): due to input corruption and factorized prediction over masked positions, it introduces a pre-train--fine-tune discrepancy and neglects dependencies among masked targets~\cite{yang2019xlnet}.

\noindent\textbf{Gradient Flow Density and Hierarchical Inductive Bias.}
We define the Direct Supervision Density (SD) as the expected fraction of tokens serving as targets per update. With mask ratio $\rho$, we have $\text{SD}_{\text{NTP}} \approx 1 \gg \text{SD}_{\text{MNP}} = \rho$.
Masked models face a dilemma: increasing $\rho$ improves density but exacerbates context corruption~\cite{wettig2023should}; for instance, KPGT relies on global nodes to stabilize optimization at $\rho=0.5$~\cite{li2023knowledge}.
In contrast, CamS achieves maximal density on uncorrupted contexts and the same structure contributes gradients repeatedly across scales, yielding substantially higher sample efficiency than sparse masked supervision~\cite{clark2020electra}.
From another perspective, Graph Transformers rely on implicit depth to propagate information. In contrast, CamS injects an \textbf{explicit structural hierarchy} via the multi-scale sequence $\mathbf{X}$, where the causal mask ensures coarse-scale predictions have deterministic access to fine-scale details. We hypothesize this bias is particularly beneficial for stabilizing predictions in activity-cliff regimes (as supported by ablation study in Section~\ref{sec:ablation}).

\section{Experiment}
\label{sec:experiment}
\subsection{Experimental Setup}
\label{sec:exp_setup}
\textbf{CamS-Tokenizer Configuration.}
We train the CamS-Tokenizer on ChEMBL-34 \cite{gaulton2017chembl} to balance diversity with feasibility, as vocabulary materialization involves expensive subgraph-isomorphism matching \cite{landrum2021rdkit, ehrlich2011maximum} that prohibits direct training on billion-scale corpora.
We learn the merge list $\mathcal{O}$ once with $K{=}685$ operations to materialize the full CamS Vocabulary ($67\mathrm{K}$).
To instantiate the multi-scale context ($M{=}4$), we slice $\mathcal{O}$ at prefix indices $k \in \{0, 62, 210, 685\}$, inducing Motif Scales $s \in \{1\mathrm{K}, 7\mathrm{K}, 27\mathrm{K}, 67\mathrm{K}\}$. These scales are selected to span the granularity spectrum: the $1\mathrm{K}$ scale serves as the atomic baseline (fine limit), the $67\mathrm{K}$ scale represents the experiment-maximal coarse limit, while $7\mathrm{K}$ and $27\mathrm{K}$ provide necessary \textbf{intermediate resolutions} to bridge the gap. Full details are in \textbf{Appendix~\ref{app:tokenizer_vocab}}.

\textbf{Pre-training Implementation.}
We instantiate a 16-layer LLaMA decoder (hidden size 720, 8 heads; $\sim$100M parameters, comparable to KPGT).
Pre-training uses Enamine675M (675M molecules)~\cite{shivanyuk2007enamine}.
Following Section~\ref{sec:method_cams_llama}, we generate five sequence views per molecule, treating them as independent instances and uniformly shuffling the resulting dataset.
We optimize the NTP objective (Eq.~\ref{eq:ntp_loss}) using hyperparameters summarized in Appendix~\ref{app:training_details}.

\textbf{Downstream Benchmarks and Baselines.}
We evaluate on MoleculeNet (11 general property tasks) \cite{wu2018moleculenet} and MoleculeACE (30 activity-cliff tasks) \cite{van2022exposing}, strictly adhering to the data splits and protocols established by the SOTA baseline KPGT \cite{li2023knowledge} (details in Appendix~\ref{app:benchmark_details}).
We benchmark against KPGT and the baselines adopted in KPGT's evaluation.
\textbf{Crucially, distinct from KPGT's heavy reliance on rich descriptor at all stages, we inject lightweight fingerprints only during fine-tuning, maintaining a purely structural pre-training}.
% On \textbf{MoleculeNet}, we include comprehensive graph-based self-supervised baselines, such as MolFormer (MolF)~\cite{wu2023molformer}, ContextPred (Ctxt)~\cite{hu2020strategies}, GROVER (GROV)~\cite{rong2020self}, JOAO~\cite{you2020graph}, GEM~\cite{fang2022geometry}, GraphMAE (GMAE)~\cite{hou2022graphmae}, and MoleBERT (MBRT)~\cite{xia2023mole} (full list in Appendix~\ref{app:detailed_results}). 
On \textbf{MoleculeNet}, we include comprehensive graph-based self-supervised baselines, such as MolFormer (MolF)~\cite{wu2023molformer}, ContextPred (Ctxt)~\cite{hu2020strategies}, GROVER (GROV)~\cite{rong2020self}, JOAO~\cite{you2020graph}, GEM~\cite{fang2022geometry}, GraphMAE (GMAE)~\cite{hou2022graphmae}, and MoleBERT (MBRT)~\cite{xia2023mole}. 
Additionally, we benchmark against the SMILES-NTP FM \textbf{NatureLM}~\cite{xia2025nature} (comparison in Appendix~\ref{app:naturelm}).
On \textbf{MoleculeACE}, we additionally include strong classical machine learning baselines (SVM, RF, GBM~\cite{van2022exposing} with ECFP (E)~\cite{rogers2010extended} or MACCS (M)~\cite{durant2002reoptimization} descriptors), which are known to outperform deep learning methods on the activity-cliff stress test.

\subsection{Results on Downstream Tasks}
\label{sec:main_results}

\textbf{General Property Prediction (MoleculeNet).}
Table~\ref{tab:MoleculeNet_final} demonstrates that CamS-LLaMA establishes a new SOTA on MoleculeNet. 
While maintaining a comparable model scale to KPGT ($\sim$100M) with fewer priors, CamS-LLaMA surpasses it in both classification (AVG-AUROC: \textbf{0.845} vs.\ 0.843) and regression (AVG-RMSE: \textbf{1.172} vs.\ 1.175).
Crucially, these results underscore superior \textbf{scalability and efficiency}: unlike KPGT's descriptor-imposed bottleneck on data scaling or \textbf{NatureLM}'s reliance on massive parameters ($\sim$56B, see Appendix~\ref{app:naturelm}), CamS leverages a \textbf{purely sequence-driven pre-training} to efficiently scale to 675M pure molecular data (discussion in Appendix~\ref{app:data_scale}) and achieve SOTA with a compact backbone.
By winning 6 out of 11 individual tasks, \textbf{CamS-LLaMA proves that properly serialized causal modeling captures molecular properties as effectively as descriptor-enhanced graph methods, without their reliance on external domain priors.} Full per-baseline results are provided in Appendix~\ref{app:detailed_results}.

\textbf{Activity Cliff Prediction (MoleculeACE).}
MoleculeACE serves as a critical stress test for a model's sensitivity to subtle structural edits.
As shown in Table~\ref{tab:moleculeace_final}, CamS-LLaMA achieves the best average RMSE of \textbf{0.624}, outperforming KPGT (0.633) by 1.4\% and the strongest ML baseline SVM$_{\text{E}}$ (0.675) by 7.6\%.
While KPGT has already demonstrated that strong pre-training can surpass ML baselines on this benchmark, CamS-LLaMA \textbf{pushes the performance boundary further with fewer external priors}.
Specifically, CamS-LLaMA ranks \textbf{1st on 17/30 tasks} (compared to KPGT's fewer wins) and achieves top-2 performance on 29/30 tasks.
This indicates that our multi-scale tokenization provides a more nuanced structural discrimination than KPGT's descriptor-based approach. 
Mechanistically, while KPGT relies on fixed descriptors that may miss non-standard structural variations driving activity cliffs, \textbf{CamS's data-driven motif mining and multi-scale serialization explicitly expose these local edits in the causal context, making them harder for the model to ignore.} Full per-baseline results are provided in Appendix~\ref{app:detailed_results}.

\textbf{Unified Takeaway.}
The results across Tables~\ref{tab:MoleculeNet_final} and \ref{tab:moleculeace_final} reveal a distinct advantage of the CamS framework: it combines \textbf{general-purpose robustness} with \textbf{fine-grained structural discriminability}.
The three-stage design—(1) motif mining, (2) causal serialization, and (3) multi-scale concatenation—allows the model to encode both global scaffolds and local variations effectively.
By doing so, CamS-LLaMA achieves what KPGT attempts via explicit descriptors: it forces the model to attend to chemically significant substructures, but does so \textbf{intrinsically} through the vocabulary and causal objective rather than \textbf{extrinsically} through handcrafted features.
This motivates the targeted interpretability analysis in Section~\ref{sec:interpretability}, where we verify whether the model indeed allocates sharper attention to cliff-differential tokens compared to baselines.

\begin{table*}[t]
\caption{Performance on MoleculeNet. Values represent Mean$_{(\text{SD})}$ over 3 random seeds. \textbf{AVG}: Average score across tasks within each task category. \textbf{Bold}: 1st; \underline{Underline}: 2nd, \textit{Italic} 3rd, \textcolor{gray}{Gray} Others. \textbf{Prev. Best}: Best baseline excluding KPGT.}
\label{tab:MoleculeNet_final}
\begin{center}
\begin{small}
\begin{sc}
% === 修复开始 ===
% 1. 将横线(booktabs rules)上下的额外间距设为0，防止右边的横线撑大左边的行距
\setlength{\aboverulesep}{0pt}
\setlength{\belowrulesep}{0pt}
% 2. 因为去掉了横线间距，稍微增加整体行高(1.2倍)，避免文字贴在线上，且使行距视觉上完全均匀
\renewcommand{\arraystretch}{1.2}
% === 修复结束 ===
\resizebox{\textwidth}{!}{
\setlength{\tabcolsep}{8pt}
\begin{tabular}{lcccclcccc}
\toprule
\multirow{2}{*}{\textbf{Task}} & \multicolumn{2}{c}{Prev. Best} & \multirow{2}{*}{KPGT} & \multirow{2}{*}{Ours} & \multirow{2}{*}{\textbf{Task}} & \multicolumn{2}{c}{Prev. Best} & \multirow{2}{*}{KPGT} & \multirow{2}{*}{Ours} \\
\cmidrule(lr){2-3} \cmidrule(lr){7-8}
 & Method & Score & & & & Method & Score & & \\
\midrule % 大横线

% === Row 1: 左边分类标题 | 右边分类(续)标题 ===
\multicolumn{5}{c}{\textit{Classification Tasks (AUROC $\uparrow$)}} & \multicolumn{5}{c}{\textit{Classification Tasks Cont.}} \\
\cmidrule(r){1-5}  \cmidrule(r){6-10}% 左右分别划线

% === Row 2: BACE | Tox21 ===
BACE & GEM & \underline{0.857}$_{(0.016)}$ & \textcolor{gray}{0.855}$_{(0.014)}$ & \textbf{0.870$_{(0.013)}$} & Tox21 & Ctxt & \underline{0.840}$_{(0.028)}$ & \textbf{0.848$_{(0.017)}$} & \textcolor{gray}{0.827}$_{(0.028)}$\\

% === Row 3: BBBP | AVG(Cls) ===
\cmidrule(l){6-10} % 只在右边画横线
BBBP & GEM & \textit{0.895$_{(0.024)}$} & \underline{0.908}$_{(0.012)}$ & \textbf{0.942$_{(0.015)}$} & \textbf{AVG (Cls)} & GEM & \textit{0.825} & \underline{0.843} & \textbf{0.845} \\
% 右边划线准备开始回归部分。由于我们设了rulesep为0，这里不会撑大行距
\cmidrule(l){6-10} 

% === Row 4: ClinTox | 右边回归标题 ===
% 这一行左边是数据，右边是标题。现在它们将完美处于同一水平线上，且上下间距与普通行一致
ClinTox & GEM & \textit{0.905$_{(0.027)}$} & \textbf{0.946$_{(0.026)}$} & \underline{0.935}$_{(0.017)}$ & \multicolumn{5}{c}{\textit{Regression Tasks (RMSE $\downarrow$)}} \\ 
\cmidrule(l){6-10} % 右边划线，回归数据开始

% === Row 5: Estrogen | ESOL ===
Estrogen & GEM & \textit{0.894$_{(0.048)}$} & \underline{0.906}$_{(0.034)}$ & \textbf{0.917$_{(0.050)}$} & ESOL & GEM & \underline{0.803}$_{(0.051)}$ & \textit{0.804$_{(0.102)}$} & \textbf{0.761$_{(0.046)}$} \\

% === Row 6: Metstab | FreeSolv ===
Metstab & GROV & \textit{0.876$_{(0.046)}$} & \underline{0.889}$_{(0.057)}$ & \textbf{0.891$_{(0.059)}$} & FreeSolv & MolF & \textit{2.322$_{(0.613)}$} & \underline{2.121}$_{(1.025)}$ & \textbf{2.110$_{(0.959)}$} \\

% === Row 7: SIDER | Lipo ===
SIDER & JOAO & \textit{0.640$_{(0.012)}$} & \underline{0.649}$_{(0.011)}$ & \textbf{0.655$_{(0.016)}$} & Lipo & GROV & \underline{0.625}$_{(0.007)}$ & \textbf{0.600$_{(0.012)}$} & \textit{0.645$_{(0.023)}$} \\

% === Row 8: ToxCast | AVG(Reg) ===
\cmidrule(l){6-10} % 只在右边画横线
ToxCast & GEM & \underline{0.733}$_{(0.020)}$ & \textbf{0.745$_{(0.003)}$} & \textit{0.724$_{(0.008)}$} & \textbf{AVG (Reg)} & MolF & \textit{1.272} & \underline{1.175} & \textbf{1.172} \\
\bottomrule
\end{tabular}
}
\end{sc}
\end{small}
\end{center}
\end{table*}

\begin{table*}[t]
\caption{Performance on MoleculeACE (RMSE $\downarrow$). \textbf{AVG}: Average RMSE across all 30 tasks. \textbf{Bold}: 1st; \underline{Underline}: 2nd; \textit{Italic} 3rd. \textbf{Prev. Best}: Best baseline excluding KPGT. \textbf{Type}: \textbf{ML} denotes traditional machine learning; \textbf{DL} denotes deep learning.}
\label{tab:moleculeace_final}
\begin{center}
\begin{small}
\begin{sc}
% 保持紧凑行距
\setlength{\aboverulesep}{0pt}
\setlength{\belowrulesep}{0pt}
\renewcommand{\arraystretch}{1.15}
\resizebox{\textwidth}{!}{
% 修改列定义：增加 Type 列
\begin{tabular}{l ccc cc @{\hskip 0.3in} l ccc cc}
\toprule
\multirow{2}{*}{\textbf{Task}} & \multicolumn{3}{c}{Prev. Best} & \multirow{2}{*}{KPGT} & \multirow{2}{*}{\textbf{Ours}} & \multirow{2}{*}{\textbf{Task}} & \multicolumn{3}{c}{Prev. Best} & \multirow{2}{*}{KPGT} & \multirow{2}{*}{\textbf{Ours}} \\
\cmidrule(lr){2-4} \cmidrule(lr){8-10} % 横线跨度改为3列
 & Method & Type & Score & & & & Method & Type & Score & & \\
\midrule % 主横线

% === Row 1: 左侧开始数据 | 右侧放标题 ===
CHEMBL1862$_{\text{Ki}}$ & GROV & DL & \textit{0.668} & \underline{0.633} & \textbf{0.600} & \multicolumn{6}{c}{\textit{Tasks Cont.}} \\
\cmidrule(r){7-12}% 右侧划线

% === Row 2-15: 正常数据行 ===
CHEMBL1871$_{\text{Ki}}$ & SVM$_{\text{E}}$ & ML & \textit{0.668} & \underline{0.605} & \textbf{0.604} & CHEMBL237$_{\text{Ki}}$ & GROV & DL & \underline{0.660} & \textit{0.678} & \textbf{0.659} \\
CHEMBL2034$_{\text{Ki}}$ & GROV & DL & \textit{0.680} & \underline{0.679} & \textbf{0.619} & CHEMBL238$_{\text{Ki}}$ & GBM$_{\text{E}}$ & ML & \textit{0.611} & \underline{0.537} & \textbf{0.537} \\
CHEMBL204$_{\text{Ki}}$ & SVM$_{\text{E}}$ & ML & \underline{0.705} & \textbf{0.666} & \textit{0.709} & CHEMBL239$_{\text{EC50}}$ & SVM$_{\text{E}}$ & ML & \textit{0.681} & \textbf{0.644} & \underline{0.647} \\
CHEMBL2047$_{\text{EC50}}$ & GMAE & DL & \underline{0.578} & \textit{0.588} & \textbf{0.519} & CHEMBL244$_{\text{Ki}}$ & GROV & DL & \textit{0.710} & \underline{0.698} & \textbf{0.696} \\
CHEMBL214$_{\text{Ki}}$ & GROV & DL & \textit{0.663} & \underline{0.652} & \textbf{0.635} & CHEMBL262$_{\text{Ki}}$ & SVM$_{\text{E}}$ & ML & \textit{0.703} & \textbf{0.627} & \underline{0.629} \\
CHEMBL2147$_{\text{Ki}}$ & SVM$_{\text{E}}$ & ML & \textit{0.612} & \underline{0.587} & \textbf{0.577} & CHEMBL264$_{\text{Ki}}$ & SVM$_{\text{E}}$ & ML & \textit{0.583} & \underline{0.574} & \textbf{0.562} \\
CHEMBL218$_{\text{EC50}}$ & RF$_{\text{M}}$ & ML & \textit{0.666} & \textbf{0.625} & \underline{0.632} & CHEMBL2835$_{\text{Ki}}$ & RF$_{\text{E}}$ & ML & \textit{0.410} & \textbf{0.373} & \underline{0.384} \\
CHEMBL219$_{\text{Ki}}$ & GROV & DL & \textit{0.737} & \textbf{0.718} & \underline{0.729} & CHEMBL287$_{\text{Ki}}$ & GROV & DL & \textit{0.732} & \underline{0.706} & \textbf{0.685} \\
CHEMBL228$_{\text{Ki}}$ & GROV & DL & \textit{0.690} & \underline{0.669} & \textbf{0.669} & CHEMBL2971$_{\text{Ki}}$ & GBM$_{\text{E}}$ & ML & \textit{0.606} & \textbf{0.571} & \underline{0.574} \\
CHEMBL231$_{\text{Ki}}$ & GROV & DL & \textit{0.649} & \textbf{0.610} & \underline{0.630} & CHEMBL3979$_{\text{EC50}}$ & GBM$_{\text{E}}$ & ML & \textit{0.686} & \underline{0.669} & \textbf{0.639} \\
CHEMBL233$_{\text{Ki}}$ & GROV & DL & \textit{0.707} & \textbf{0.691} & \underline{0.692} & CHEMBL4005$_{\text{Ki}}$ & SVM$_{\text{E}}$ & ML & \underline{0.550} & \textit{0.559} & \textbf{0.543} \\
CHEMBL234$_{\text{Ki}}$ & SVM$_{\text{E}}$ & ML & \textit{0.637} & \textbf{0.606} & \underline{0.624} & CHEMBL4203$_{\text{Ki}}$ & MBRT & DL & \underline{0.820} & \textit{0.830} & \textbf{0.787} \\
CHEMBL235$_{\text{EC50}}$ & RF$_{\text{E}}$ & ML & \textit{0.637} & \underline{0.624} & \textbf{0.612} & CHEMBL4616$_{\text{EC50}}$ & SVM$_{\text{E}}$ & ML & \textit{0.589} & \underline{0.587} & \textbf{0.538} \\
CHEMBL236$_{\text{Ki}}$ & SVM$_{\text{E}}$ & ML & \textit{0.692} & \textbf{0.655} & \underline{0.669} & CHEMBL4792$_{\text{Ki}}$ & SVM$_{\text{E}}$ & ML & \textit{0.675} & \textbf{0.619} & \underline{0.651} \\

% === Row 16: 左侧第16个任务 | 右侧 AVG ===
\cmidrule(l){7-12} % 只在右边画横线
CHEMBL237$_{\text{EC50}}$ & SVM$_{\text{E}}$ & ML & \textit{0.760} & \underline{0.716} & \textbf{0.684} & \textbf{AVG (Overall)} & SVM$_{\text{E}}$ & ML & \textit{0.675} & \underline{0.633} & \textbf{0.624} \\
\bottomrule
\end{tabular}
}
\end{sc}
\end{small}
\end{center}
\end{table*}

\subsection{Interpretability: Attention on Activity Cliffs}
\label{sec:interpretability}
\textbf{Interpretability Setup and Metric.}
To mechanistically explain the activity-cliff superiority observed in Table~\ref{tab:moleculeace_final}, we investigate whether the CamS-LLaMA attention mechanism inherently prioritizes cliff-driving structural variations.
Aligned with the CamS-Tokenizer framework (Section~\ref{sec:method_tokenizer}), we map atom-level differences in activity-cliff pairs to tokens within specific \textbf{Motif Scales} of the concatenated CamS Sequence $\mathbf{X}$.
Specifically, for each cliff pair, we identify differential versus shared atoms and project these labels onto the CamS Sequence $\mathbf{X}$.
Using the final-layer attention distribution, we compute the Mean Attention on Differential Tokens (MDTA) and Shared Tokens (MSTA) within each scale region $s$. We define the \textbf{Relative Differential-Token Attention Preference (Rel-DTAP)} as:
\begin{equation}
\label{eq:rel_dtap}
\mathrm{Rel\text{-}DTAP}_s = \frac{\mathrm{MDTA}_s-\mathrm{MSTA}_s}{\mathrm{MSTA}_s+\epsilon}\times 100 \%,
\end{equation}
averaged over all test pairs. A positive value indicates that the model allocates disproportionately higher attention to the subtle structural edits driving the activity cliff.
Full implementation details are in Appendix~\ref{app:interpret}.

\textbf{Findings and Implications.}
Table~\ref{tab:interpretability_rel_dtap} reveals three patterns that corroborate our design principles:
(1) \textbf{Intrinsic Structural Discriminability}: On the full sequence, Rel-DTAP is consistently positive ($\sim$14\%), confirming that even without explicit guidance, the NTP-driven causal objective naturally steers attention toward cliff-driving motifs.
(2) \textbf{Scale-Dependent Hierarchy}: Coarse scales (27K/67K) exhibit significantly stronger preference ($\sim$25\%) than the fine-grained baseline (1K, $\sim$8\%). This validates our hypothesis in Section~\ref{sec:method_tokenizer} that coarse motifs act as high-level "semantic anchors," making structural edits more salient compared to the fragmented 1K view.
(3) \textbf{Fingerprint as a Stabilizer}: Fingerprint injection (Section~\ref{sec:method_cams_llama}) does not uniformly boost attention but acts as a corrective stabilizer. It specifically rectifies the anomalous attention pattern at the intermediate 7K scale (from -0.18\% to 2.10\%), resolving ambiguity where motif granularity may be suboptimal.

Collectively, these findings provide a mechanistic basis for CamS's dominance on MoleculeACE: \textbf{the multi-scale architecture explicitly exposes subtle edits as distinct multi-granular tokens, preventing them from being obscured by the local averaging mechanisms common in standard graph encoders.}

\begin{table}[t]
\caption{\textbf{Relative Differential-Token Attention Preference (Rel-DTAP).}
We report the average Rel-DTAP across all activity-cliff test pairs.
A positive value indicates that the model attends more intensively to \textbf{differential tokens}.
\textbf{Scale Region ($s$)}: Indicates the model's attention to regions in CamS Sequence corresponding to specific Motif Scales.\textbf{With FP / Without FP}: Indicates whether the fingerprint was injected.}
\label{tab:interpretability_rel_dtap}
\vskip 0.1in
\begin{center}
\begin{small}
\setlength{\tabcolsep}{6pt}
\begin{tabular}{lcc}
\toprule
Scale Region ($s$) & Without FP & With FP \\
\midrule
1K (Fine)   & 8.08\%  & 4.72\% \\
7K          & -0.18\% & 2.10\% \\
27K         & 24.33\% & 23.60\% \\
67K (Coarse)& 25.87\% & 24.00\% \\
\midrule
CONCAT (All) & 14.79\% & 14.23\% \\
\bottomrule
\end{tabular}
\end{small}
\end{center}
\vskip -0.1in
\end{table}

\begin{figure}[ht]
\vskip 0.1in
\begin{center}
\centerline{\includegraphics[width=\columnwidth]{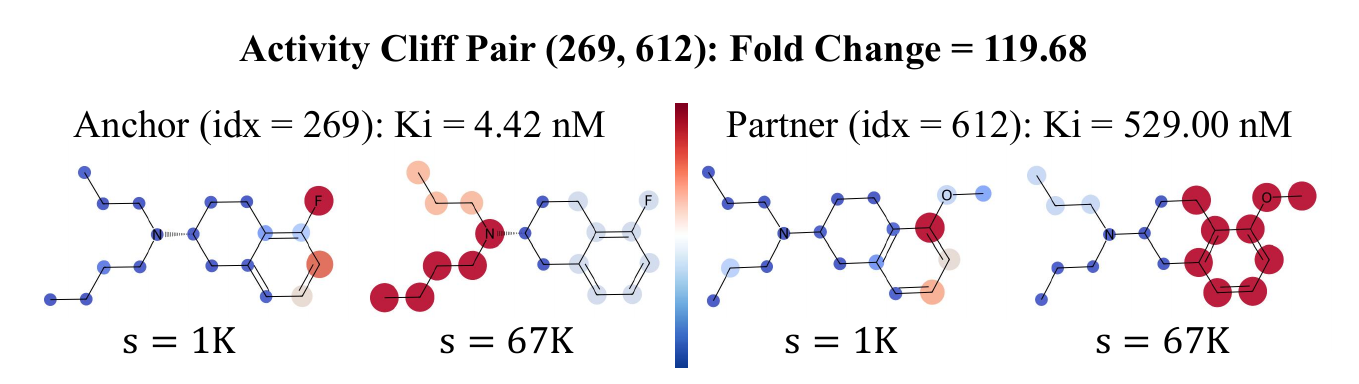}}
\caption{\textbf{Attention Visualization on an Activity Cliff Pair (CHEMBL234$_{\text{Ki}}$).}
Attention heatmaps for the Anchor and Partner molecules at Motif Scales $s=1\mathrm{K}$ and $s=67\mathrm{K}$. Nodes are colored by their attention weights (Red: High, Blue: Low).}
\label{fig:case_study}
\end{center}
\vskip -0.1in
\end{figure}

\textbf{Case Study.}
Figure~\ref{fig:case_study} visualizes the attention landscape for a representative activity-cliff pair from CHEMBL234$_{\text{Ki}}$ (Fold Change $\approx 120$).
The anchor and partner molecules differ in two subtle aspects: a substitution on the benzene ring (fluorine vs.\ methoxy group) and a variation in the lower-left chain structure.
We visualize the attention weights corresponding to these molecules across the two extreme scales ($s=1\mathrm{K}$ and $s=67\mathrm{K}$) within the unified CamS Sequence. We observe two distinct attention patterns that highlight structural discriminability:
(1) \textbf{Motif-Level Amplification at Coarse Scales ($67\mathrm{K}$):} The cliff-driving substitutions result in the emergence of entirely different high-level motif tokens at the $67\mathrm{K}$ scale. The model effectively ``isolates'' the activity shift by allocating intense attention (red nodes) specifically to these differential motifs, treating them as salient semantic anchors.
(2) \textbf{Cross-Scale Consistency at Fine Scales ($1\mathrm{K}$):} Even at the fragmented atomic baseline ($1\mathrm{K}$), , attention is not uniformly distributed but concentrates near the modification sites. This confirms that the causal information flow from fine-to-coarse scales enables the model to pinpoint local edits, back-propagating relevance from high-level motifs to their constituent atoms.
This visual evidence reinforces our statistical findings: CamS does not merely memorize structures but actively \textbf{attends to and discriminates} the precise sub-structural variations that govern molecular potency.
Case selection rules with additional case studies are provided in Appendix~\ref{app:case_selection} and~\ref{app:more_cases}.

\subsection{Ablation Study}
\label{sec:ablation}
\textbf{Ablation setup.}
We evaluate three variants against the full model (summary in Table~\ref{tab:ablation_rank}, full breakdown in Appendix~\ref{app:ablation_details}) to isolate core components: (1) \textbf{w/o FP} completely removes fingerprint injection; (2) \textbf{$1\mathrm{K}$ Only} and (3) \textbf{$67\mathrm{K}$ Only} use single fixed scales. Single-scale variants have pre-training and fine-tuning budgets matched to the full model.

\paragraph{Indispensability of Multi-Scale Context.}
The full model consistently outperforms single-scale variants (e.g., MolACE 0.624 vs.\ 0.649), confirming that Cross-Scale Concatenation (Section~\ref{sec:method_tokenizer}) is fundamental.
By enabling coarse reasoning to ground on fine evidence, this design validates our central thesis: multi-scale causal serialization is the key representation-level innovation that unlocks vanilla NTP for molecular graphs modeling.

\paragraph{Pitfall of Coarse-Scale Over-Compression.}
Single-scale results show that coarse modeling ($s=67\mathrm{K}$) consistently lags behind fine-grained modeling ($s=1\mathrm{K}$).
We attribute this to \textbf{supervision sparsity}: over-compressed sequences provide insufficient AR steps for effective learning.
The full model circumvents this by concatenation, successfully fusing the dense supervision of fine scales with the coarse-scale scaffold-level abstraction, without committing to a specific ``best'' resolution.

\paragraph{Fingerprint as the Maximum-Scale Global Token.}
The pure-sequence model (\textit{w/o FP}) proves remarkably robust, particularly in classification where it statistically rivals KPGT (0.838 vs.\ 0.843).
While KPGT resorts to heavy priors to patch MNP's limitations (e.g., over-smoothing), our use of the fingerprint is fundamentally different: we strategically integrate it as the \textbf{maximum-scale global token} (equivalent to Motif Scale $s = \infty$) to complete the topological hierarchy.
Acting as a stabilizer (Section~\ref{sec:interpretability}), it plays a significantly larger role in regression tasks, effectively completing the final link of our multi-scale structural context.
Thus, the full model's superiority stems from \textbf{a unified causal formulation:} \textbf{$Global + (Fine \to Coarse)$}, where the fingerprint serves as the holistic structural anchor driving the fine-to-coarse reasoning.
\begin{table}[t]
\caption{\textbf{Ablation study}. Scores represent the average performance across all tasks in the respective benchmarks from Tables \ref{tab:MoleculeNet_final} and \ref{tab:moleculeace_final}. \textbf{R}: Global ranking in complete baselines.}
\label{tab:ablation_rank}
\vskip 0.1in
\begin{center}
\begin{small}
\begin{sc}
\setlength{\tabcolsep}{2.5pt} 
\resizebox{\columnwidth}{!}{
\begin{tabular}{lcccccc}
\toprule
\multirow{2}{*}{Method} & \multicolumn{2}{c}{MolNet-Cls} & \multicolumn{2}{c}{MolNet-Reg} & \multicolumn{2}{c}{MolACE} \\
\cmidrule(lr){2-3} \cmidrule(lr){4-5} \cmidrule(lr){6-7}
 & AUROC ($\uparrow$) & R & RMSE ($\downarrow$) & R & RMSE ($\downarrow$) & R \\
\midrule
% --- Our Work ---
\textbf{CamS-LLaMA} & \textbf{0.845} & 1 & \textbf{1.172} & 1 & \textbf{0.624} & 1 \\
\quad -- w/o FP & 0.838 & 3 & 1.195 & 3 & 0.650 & 5 \\
\quad -- 1K Only & 0.833 & 4 & 1.215 & 4 & 0.641 & 3 \\
\quad -- 67K Only & 0.818 & 6 & 1.329 & 7 & 0.649 & 4 \\
\midrule
% --- Baselines ---
KPGT & 0.843 & 2 & 1.175 & 2 & 0.633 & 2 \\
Prev. Best & 0.825 (GEM) & 5 & 1.272 (MolF) & 5 & 0.675 (SVM$_{\text{E}}$) & 6 \\
\bottomrule
\end{tabular}
}
\end{sc}
\end{small}
\end{center}
\vskip -0.15in
\end{table}

\section{Conclusion}
\label{sec:conclusion}
We presented \textbf{CamS}, a graph-to-sequence interface that enables standard decoder-only Transformers to learn molecular topology via Next Token Prediction. Its FM prototype \textbf{CamS-LLaMA} achieve SOTA on MoleculeNet and MoleculeACE. Our multi-scale causal serialization is fundamental to this success, as mechanistic analysis confirms it explicitly drives attention toward cliff-driving structural differences. Ultimately, \textbf{CamS validates generic AR FMs as powerful engines for structure-native molecular science given the right representation interface.}

%%%%%%%%%
\bibliography{example_paper}
\bibliographystyle{icml2026}
%%%%%%%%%

\newpage
\appendix
\onecolumn

% =========================================================
% Appendix A: Tokenizer & Serialization Details (Method refs)
% =========================================================
\section{CamS-Tokenizer and Graph-to-Sequence Construction}
\label{app:tokenizer_details}
\textbf{Overview.} This section supplements the CamS-Tokenizer framework description in Sec.~\ref{sec:method_tokenizer} and the CamS-LLaMA pipeline in Sec.~\ref{sec:method_cams_llama}. It provides implementation-level details for:
(1) \textbf{Vocabulary Mining}: Learning merge operations and constructing the Single-Atom Vocabulary Closure (SAVC) (Algs.~\ref{alg:merge_learning} and \ref{alg:savc});
(2) \textbf{Materialization}: Building the connection-aware motif vocabulary (Alg.~\ref{alg:vocab_materialize});
(3) \textbf{Per-Scale Encoding}: The recursive encoding process with unknown recovery (Alg.~\ref{alg:encode_split}) and the deterministic Scaffold-Rooted BFS serialization (Alg.~\ref{alg:motif_bfs_order});
(4) \textbf{Cross-Scale Concatenation}: Constructing multi-view training instances for NTP (Alg.~\ref{alg:multiscale_concat}).

\subsection{Tokenizer Vocabulary Mining and Single-Atom Coverage}
\label{app:tokenizer_vocab}

\paragraph{Notation and Objects.}
We start from an RDKit molecule and construct the atom graph $G_0=(V_0,E_0)$, where nodes are atoms and edges are chemical bonds.
A merge list is an ordered sequence of operations $\mathcal{O}=(o_1,\dots,o_K)$, where each operation $o_t$ is a canonical fragment code extracted from the union of two adjacent nodes.
The \textbf{Motif Scale} $s$ corresponds to a prefix length $k_s$, such that applying $\mathcal{O}_{\le k_s}$ to $G_0$ yields a motif graph $G_s=(V_s,E_s)$, where each node $v\in V_s$ covers a subset of atoms $\mathrm{atom\_indices}(v)\subseteq V_0$.

\paragraph{Connection-Aware Motif Representation.}
Each motif token is stored in a \emph{pair form} $(v_{\mathrm{noConn}}, v_{\mathrm{withConn}})$.
To ensure a deterministic and unique identifier for every substructure, we employ RDKit's canonicalization routine \cite{landrum2021rdkit}.
The first component, $v_{\mathrm{noConn}}$, is the standard canonical SMILES of the fragment.
The second component, $v_{\mathrm{withConn}}$, explicitly encodes connectivity information to differentiate chemically identical fragments with different attachment contexts (e.g., a pyridine ring attached at the 2-position vs.\ the 3-position).
Specifically, at each attachment site (severed bond), we insert a dummy atom (wildcard \texttt{*}) preserving the original bond type. 
To guarantee a unique string representation invariant to atom ordering, we generate the canonical SMILES of this wildcard-augmented fragment.
Thus, two motifs are identical if and only if they share the same graph topology, atom types, and attachment configurations.
We materialize these pairs by replaying $\mathcal{O}_{\le k_s}$ on the tokenizer corpus (Alg.~\ref{alg:vocab_materialize}).

\paragraph{Single-Atom Vocabulary Closure (SAVC).}
A key challenge for predictive encoding is \emph{coverage}: if a rare ``single-atom + attachment-pattern'' form is absent from the mined vocabulary, standard tokenizers collapse it to \texttt{[UNK]}, creating causal information breakpoints.
SAVC addresses this by: (1) enumerating common connection-aware single-atom motifs for all typical valences, and (2) introducing an element-wise fallback token \texttt{[X\_AltForm]} to represent rare/atypical forms.
During encoding, if a queried motif contains exactly one core atom of element $X$ but the exact form is missing, we back off to \texttt{[X\_AltForm]} instead of \texttt{[UNK]} (Alg.~\ref{alg:savc}).

\paragraph{Encoding-Time Unknown Recovery.}
Encoding greedily applies $\mathcal{O}_{\le k_s}$ and maps resulting motifs to token IDs.
If a motif is unknown (not in $\Sigma$), we recursively split it along its stored merge tree until known sub-motifs are obtained.
The recursion guarantees termination at single-atom leaves, where SAVC back-off logic applies (Alg.~\ref{alg:encode_split}).

\subsection{Graph-to-Causal-Sequence Serialization}
\label{app:graph2seq}

\paragraph{Motif Graph Construction.}
After applying $\mathcal{O}_{\le k_s}$ to $G_0$, we obtain the motif graph $G_s=(V_s,E_s)$. An edge $(u,v)\in E_s$ exists if any original bond in $G_0$ connects atoms across fragments $u$ and $v$.

\paragraph{Scaffold-Rooted BFS Order (Intra-Scale Order).}
CamS serializes $G_s$ into a sequence via a deterministic \textbf{Scaffold-Rooted BFS}.
To ensure that the serialization is invariant to the input atom permutation (i.e., canonical), we perform a standard molecule-level canonicalization (via RDKit) \textit{before} any processing.
This establishes a canonical ordering of the original atom indices $0, \dots, N-1$.
During BFS on the motif graph $G_s$, we define the \texttt{node\_id} of a motif $v$ as the minimum canonical atom index among its constituent atoms ($\min_{a \in \mathrm{atoms}(v)} \mathrm{idx}(a)$).
We then select the root as the motif with the largest atom count (breaking ties by the smallest \texttt{node\_id}) and traverse $G_s$ breadth-first.
When expanding neighbors, we sort them by their \texttt{node\_id} in ascending order.
This strategy prioritizes global backbone structure before local substituents, establishing a stable \textbf{Center-to-Periphery} causal order (Alg.~\ref{alg:motif_bfs_order}).

\subsection{Multi-Scale Concatenation and Training Views}
\label{app:multiscale_concat}

\paragraph{Scale Definition.}
We define a set of Motif Scales $S = \{s_1, s_2, \dots, s_M\}$ (e.g., $1\mathrm{K}, 7\mathrm{K}, \dots$), each corresponding to a prefix $\mathcal{O}_{\le s_i}$.
This ``train once, slice many'' strategy allows reusing the same learned merge statistics across all scales.

\paragraph{Data Augmentation via Views.}
For each molecule, we generate $M$ single-scale views $\{X^{(s_1)}, \dots, X^{(s_M)}\}$ and one concatenated multi-scale view $\mathbf{X}$.
The multi-scale view $\mathbf{X}$ is constructed by concatenating single-scale sequences in \textbf{Fine-to-Coarse Order} (Alg.~\ref{alg:multiscale_concat}), enabling the model to leverage fine-grained details as context for coarse-grained predictions.

\paragraph{NTP Loss Masking.}
For AR training, we compute loss on all motif tokens but exclude special tokens \texttt{[BOS]}, \texttt{[EOS]}, and \texttt{[CONCAT]} from prediction targets (typically by setting labels to \texttt{-100}).

% ================= ALGORITHMS =================

% Algorithm 1: Merge Learning
\begin{algorithm}[t]
\caption{BPE-style Merge Operation Learning (List $\mathcal{O}$)}
\label{alg:merge_learning}
\begin{algorithmic}[1]
\STATE \textbf{Input:} Tokenizer corpus $\mathcal{D}$, max iterations $K$, min frequency $f_{\min}$
\STATE \textbf{Output:} Ordered merge list $\mathcal{O}=(o_1,\dots,o_{K'})$
\STATE Initialize graph $G_m$ for each molecule $m\in\mathcal{D}$ (every atom is a node)
\STATE Compute initial pair statistics for all adjacent edges
\FOR{$t=1$ \textbf{to} $K$}
    \STATE Select operation $o_t \leftarrow \arg\max_{c}\;(\mathrm{stats}[c],\,c)$
    \IF{$\mathrm{stats}[o_t] < f_{\min}$}
        \STATE \textbf{break}
    \ENDIF
    \STATE Append $o_t$ to $\mathcal{O}$
    \STATE Apply $o_t$ to all applicable edges in $\mathcal{D}$, merging nodes into larger motifs
    \STATE \emph{(Implementation Note: Merges are performed greedily. In case of overlaps, edges with smaller canonical atom indices are prioritized to ensure determinism.)}
    \STATE Update local pair statistics around merged nodes
    \STATE Reset $\mathrm{stats}[o_t]\leftarrow 0$
\ENDFOR
\end{algorithmic}
\end{algorithm}

% Algorithm 2: SAVC
\begin{algorithm}[t]
\caption{Single-Atom Vocabulary Closure (SAVC) \& Back-off}
\label{alg:savc}
\begin{algorithmic}[1]
\STATE \textbf{Input:} Element set $\mathcal{E}$, valences $\mathrm{Val}(X)$, bond types $\mathcal{B}$
\STATE \textbf{Output:} Basic single-atom vocabulary $\Sigma_{\mathrm{atom}}$
\FOR{each element $X\in\mathcal{E}$}
    \STATE Add \texttt{[X]} (standalone) and \texttt{[X\_AltForm]} (fallback) to $\Sigma_{\mathrm{atom}}$
    \FOR{each valence state and attachment pattern}
        \STATE Construct connection-aware SMILES with wildcards (e.g., \texttt{*X(*)*})
        \STATE Add to $\Sigma_{\mathrm{atom}}$
    \ENDFOR
\ENDFOR
\STATE \textbf{Encoding-Time Back-off Logic:}
\IF{queried motif is unknown AND contains exactly 1 core atom (element $X$)}
    \STATE return ID of \texttt{[X\_AltForm]}
\ELSE
    \STATE return \texttt{[UNK]}
\ENDIF
\end{algorithmic}
\end{algorithm}

% Algorithm 3: Materialization
\begin{algorithm}[t]
\caption{Connection-Aware Vocabulary Materialization}
\label{alg:vocab_materialize}
\begin{algorithmic}[1]
\STATE \textbf{Input:} Corpus $\mathcal{D}$, operation prefix $\mathcal{O}_{\le k}$, SAVC vocabulary $\Sigma_{\mathrm{atom}}$
\STATE \textbf{Output:} Full vocabulary $\Sigma$ (Map: Canonical SMILES $\to$ ID)
\FOR{each molecule $m\in\mathcal{D}$}
    \STATE Apply $\mathcal{O}_{\le k}$ to partition atoms into motif nodes $V$
    \FOR{each motif node $v \in V$}
        \STATE Extract core fragment SMILES $v_{\mathrm{noConn}}$
        \STATE Extract connection-aware SMILES $v_{\mathrm{withConn}}$ (inserting \texttt{*} at cut bonds)
        \STATE Add $(v_{\mathrm{noConn}}, v_{\mathrm{withConn}})$ to candidate set
    \ENDFOR
\ENDFOR
\STATE Filter candidates by frequency; Union with $\Sigma_{\mathrm{atom}}$ and Special Tokens to form $\Sigma$
\end{algorithmic}
\end{algorithm}

% % Algorithm 4: Encoding & Split
% \begin{algorithm}[H]
% \caption{Recursive Encoding with Unknown Recovery}
% \label{alg:encode_split}
% \begin{algorithmic}[1]
% \STATE \textbf{Input:} Molecule $M$, operations $\mathcal{O}_{\le k}$, Vocabulary $\Sigma$
% \STATE \textbf{Output:} Ordered token list
% \STATE \textbf{Step 1: Partition.} Apply $\mathcal{O}_{\le k}$ to $M$ to build BPEGraph $G_s$.
% \STATE \textbf{Step 2: Serialize.} Order nodes $V_s$ via Scaffold-Rooted BFS (Alg.~\ref{alg:motif_bfs_order}).
% \STATE \textbf{Step 3: Tokenize with Recovery.}
% \FOR{each node $v$ in BFS order}
%     \IF{$v \in \Sigma$}
%         \STATE Emit ID($v$)
%     \ELSE
%         \STATE \textbf{Recursive Split:} Undo the last merge operation for $v$ to get children $(c_1, c_2)$
%         \STATE Recursively process $c_1$ and $c_2$
%         \STATE (Base case: if single-atom child is still unknown, apply SAVC Back-off)
%     \ENDIF
% \ENDFOR
% \end{algorithmic}
% \end{algorithm}

% Algorithm 4: Recursive Encoding with Unknown Recovery
% Matches code: graph_bpe_encode.py
\begin{algorithm}[t]
\caption{Encoding with Recursive Unknown Recovery}
\label{alg:encode_split}
\begin{algorithmic}[1]
\STATE \textbf{Input:} Molecule $M$, operations $\mathcal{O}_{\le k}$, Vocabulary $\Sigma$
\STATE \textbf{Output:} Ordered token list

\STATE \textbf{// Phase 1: Construction (Over-merge)}
\STATE Initialize atom graph $G$. Assign each node a leaf \texttt{BPETreeNode}.
\FOR{operation $o \in \mathcal{O}_{\le k}$}
    \STATE Find all edges matching $o$.
    \FOR{edge $(u, v)$ in matches (ordered by node index)}
        \STATE Merge $u, v$ into $w$.
        \STATE Record merge history: $w.\text{tree} \leftarrow \text{Node}(\text{children}=[u.\text{tree}, v.\text{tree}])$.
    \ENDFOR
\ENDFOR

\STATE \textbf{// Phase 2: Serialization \& Recovery}
\STATE Serialize nodes $V_s$ via Scaffold-Rooted BFS (Alg.~\ref{alg:motif_bfs_order}).
\STATE Initialize empty list $L$.
\FOR{node $v$ in BFS order}
    \STATE $L.\text{append}(\text{RecursiveResolve}(v, \Sigma))$
\ENDFOR
\STATE \textbf{return} $L$

\STATE \textbf{Function} \text{RecursiveResolve}(node $v$, Vocabulary $\Sigma$):
\IF{$v \in \Sigma$}
    \STATE \textbf{return} [ID($v$)]
\ELSE
    \STATE \emph{// Unknown motif: backtrack using the BPE tree built in Phase 1}
    \STATE Let $c_1, c_2 \leftarrow v.\text{tree}.\text{children}$
    \STATE \textbf{return} \text{RecursiveResolve}($c_1, \Sigma$) + \text{RecursiveResolve}($c_2, \Sigma$)
    \STATE \emph{// Base case: if single atom is unknown, apply SAVC back-off (Alg.~\ref{alg:savc})}
\ENDIF
\end{algorithmic}
\end{algorithm}

% Algorithm 5: BFS
\begin{algorithm}[t]
\caption{Scaffold-Rooted BFS (Intra-Scale Order)}
\label{alg:motif_bfs_order}
\begin{algorithmic}[1]
\STATE \textbf{Input:} Motif Graph $G_s=(V_s, E_s)$
\STATE \textbf{Output:} Ordered list of nodes
\STATE Select root $r \leftarrow \arg\max_{v\in V_s} (\text{atom\_count}(v), -\text{node\_id}(v))$
\STATE Initialize Queue $Q \leftarrow [r]$, Visited $\leftarrow \{r\}$, Order $\leftarrow []$
\WHILE{$Q$ is not empty}
    \STATE $u \leftarrow Q.\text{pop\_front}()$
    \STATE Append $u$ to Order
    \STATE Get neighbors $\mathcal{N}(u)$, sort by node\_id
    \FOR{$v \in \mathcal{N}(u)$}
        \IF{$v \notin \text{Visited}$}
            \STATE $Q.\text{push\_back}(v)$, Visited.add($v$)
        \ENDIF
    \ENDFOR
\ENDWHILE
\STATE \textbf{return} Order
\end{algorithmic}
\end{algorithm}

% Algorithm 6: Multi-scale
\begin{algorithm}[t]
\caption{Multi-Scale Concatenation (Inter-Scale Order)}
\label{alg:multiscale_concat}
\begin{algorithmic}[1]
\STATE \textbf{Input:} Molecule $M$, scales $\{s_1, \dots, s_M\}$ (Fine $\to$ Coarse)
\STATE \textbf{Output:} Single-scale views $\{X^{(s_j)}\}$, Multi-scale view $\mathbf{X}$
\FOR{$j=1$ \textbf{to} $M$}
    \STATE $X^{(s_j)} \leftarrow \text{Encode}(M, \mathcal{O}_{\le s_j})$ (using Alg.~\ref{alg:encode_split})
\ENDFOR
\STATE \textbf{Construct Concatenated View:}
\STATE $\mathbf{X} \leftarrow [\texttt{BOS}]$
\FOR{$j=1$ \textbf{to} $M$}
    \STATE Append tokens of $X^{(s_j)}$ (excluding its own BOS/EOS)
    \IF{$j < M$}
        \STATE Append \texttt{[CONCAT]}
    \ENDIF
\ENDFOR
\STATE Append \texttt{[EOS]}
\STATE \textbf{return} $\{X^{(s_1)}, \dots, X^{(s_M)}, \mathbf{X}\}$
\end{algorithmic}
\end{algorithm}

\FloatBarrier
\clearpage

% =========================================================
% Appendix B: Theory 
% =========================================================
\section{Theoretical Derivations}
\label{app:theory}
\textbf{Overview.} This section corresponds to the theoretical analysis in Sec.~\ref{sec:method_analysis}. It supplements the main text with derivation-level details of:
(1) The DPI-based proof for the information loss induced by stochastic corruption, including a discussion on Graph-specific evidence uncertainty (Sec.~\ref{app:proof_info});
(2) The quantitative analysis of Direct Supervision Density, contrasting NTP vs.\ MNP and quantifying the gain from multi-view augmentation (Sec.~\ref{app:gradient_density});
(3) The explicit structural-bias formulation that contrasts graph-transformer static bias with CamS causal constraint plus learned aggregation (Sec.~\ref{app:topology_comparison}).

\subsection{Proof of Proposition~\ref{prop:info_gain} (Context Information)}
\label{app:proof_info}

We formalize the conditioning-information gap between (1) uncorrupted evidence and (2) randomly masked evidence.

\paragraph{Setup and Markov Chain.}
For predicting token $x_t$, let $Z_t$ denote an \textbf{unmasked} evidence set (e.g., the full causal history in CamS) and let $\tilde{Z}_t=\mathcal{M}(Z_t)$ be its stochastically masked version under a masking channel $\mathcal{M}$ (e.g., random node/substructure masking used in MLM/MNP). This yields a Markov chain:
\begin{equation}
x_t \;\longleftrightarrow\; Z_t \;\longrightarrow\; \tilde{Z}_t .
\end{equation}

\begin{proof}
By the Data Processing Inequality (DPI), for any Markov chain $X\!\leftrightarrow\!Z\!\rightarrow\!\tilde{Z}$, the data processing step $\tilde{Z}$ cannot increase the mutual information with the target $X$. Formally:
\begin{equation}
I(x_t; Z_t) \ge I(x_t; \tilde{Z}_t).
\end{equation}
This proves Proposition~\ref{prop:info_gain}.
\end{proof}

\paragraph{Remark1: MLM limitations in NLP.}
Our analysis aligns with classic critiques of Masked Language Modeling (MLM). XLNet~\cite{yang2019xlnet} points out two intrinsic issues of corruption-based objectives: (1) the special token corruption (e.g., \texttt{[MASK]}) creates a pretrain-finetune discrepancy, and (2) predicting multiple masked positions with a factorized objective effectively neglects their conditional dependencies given the unmasked context.
Relatedly, ELECTRA~\cite{clark2020electra} shows that much of the compute/sample-efficiency gap of MLM stems from defining the loss only on a small masked subset rather than all positions.

\paragraph{Remark 2: Graph-Specific Evidence Instability.}
While the DPI inequality holds generally, the loss is particularly severe on graphs due to \textbf{Evidence-Pattern Uncertainty}. The most predictive evidence for a token often lies in its immediate local neighborhood (e.g., bond context, ring connectivity, functional-group surroundings).
Random masking $\mathcal{M}$ frequently removes not only the target token but also its critical neighborhood evidence ("co-masked"), making the conditional distribution $P(x_t | \tilde{Z}_t)$ highly unstable across different masking patterns.
This motivates a line of graph-pretraining designs that explicitly trade masking density for context stability:
\begin{itemize}
    \item \textbf{KPGT}~\cite{li2023knowledge}: Masks structured units and adds global knowledge nodes to anchor the context.
    \item \textbf{GROVER}~\cite{rong2020self}: Masks local subgraphs and predicts contextual properties.
    \item \textbf{StructMAE}~\cite{liu2024mask}: Uses structure-guided / curriculum masking rather than purely random masking.
\end{itemize}
These methods can be interpreted as efforts to find better operating points on the fundamental corruption--prediction trade-off of masked modeling~\cite{wettig2023should}, but they still fundamentally operate on corrupted inputs.
\textbf{CamS Implication:} In contrast, CamS conditions coarse-scale predictions on a \textbf{structurally complete} and uncorrupted fine-grained history. This ensures evidence stability and maximizes information gain without requiring auxiliary anchors.

\subsection{Direct Supervision Density Analysis (Decomposition)}
\label{app:gradient_density}

As a heuristic proxy for the richness of the learning signal, we compare the density of \textbf{direct} token-level supervision signals. To address the distinction between objective efficiency and data augmentation, we decompose the density gain into two factors: \textbf{Intrinsic Efficiency} and \textbf{Systemic Multiplier}.

\paragraph{Gradient Origins.}
Let $\mathcal{L}$ be the token-level loss.
\begin{itemize}
    \item \textbf{MNP:} Direct loss terms (hence gradient sources) occur only on masked indices $i\in\mathcal{M}$; unmasked tokens receive gradients only indirectly via attention coupling.
    \item \textbf{NTP:} Direct loss terms occur at (almost) all positions $t\in\{1,\dots,T_{\text{all}}-1\}$.
\end{itemize}

\paragraph{Factor 1: Intrinsic Objective Efficiency ($\times 1/\rho$).}
We define the \textbf{Direct Supervision Density} (SD) as the expected fraction of tokens that serve as prediction targets per update:
\begin{equation}
\text{SD}_{\text{NTP}} \approx 1,\qquad
\text{SD}_{\text{MNP}}=\rho.
\end{equation}
Thus, per pass, NTP provides about $1/\rho$ times more direct targets than MNP. For a typical masking rate $\rho=0.15$, this is a factor of $6.7\times$. This advantage is inherent to the NTP objective.

\paragraph{Factor 2: Systemic Augmentation Multiplier ($\times M$).}
Crucially, the sequence representation of CamS naturally supports lightweight multi-view augmentation. We utilize 5 views per molecule (4 single-scale + 1 multi-scale) during pre-training.
Let $\mathcal{R}$ be the total number of supervision targets provided by one molecule in one epoch:
\begin{align}
    \mathcal{R}_{\text{MNP}} &\approx 1 \times (\rho \cdot T_{\text{avg}}) \\
    \mathcal{R}_{\text{CamS}} &\approx 5 \times (1 \cdot T_{\text{avg}})
\end{align}
Even against aggressive masking ($\rho=0.5$ as in KPGT), CamS provides $10\times$ more signals per molecule. While MNP can ostensibly adopt augmentation, re-encoding massive graphs for every view is computationally costlier than re-slicing sequences.
CamS leverages the product of these factors, resulting in substantially higher sample efficiency.

\paragraph{Trade-off and Practical Masking Rates.}
Increasing $\rho$ increases the number of predictions but also increases corruption, forming a fundamental trade-off~\cite{wettig2023should}. KPGT reports that setting $\rho=0.5$ is only viable when global knowledge tokens help stabilize the conditioning context~\cite{li2023knowledge}. CamS bypasses this dilemma entirely, achieving maximum density ($\text{SD}\approx 1$) with zero corruption.

\subsection{Structural Bias Formulation}
\label{app:topology_comparison}

We distinguish the structural injection mechanism in Eq.~\eqref{eq:unified_prop} based on the bias term $\psi_{ij}$.

\paragraph{Graph Transformer (Hard Static Bias).}
Standard Graph Transformers inject structure via static encodings, such as Shortest Path Distance (SPD), added to the attention scores~\cite{ying2021transformers}:
\begin{equation}
    \psi_{ij}^{\text{Graph}} = b_{\text{SPD}(i,j)}.
\end{equation}
This imposes an isotropic prior: all atoms at distance $k$ are treated equally by the structural bias, regardless of their semantic content.

\paragraph{CamS (Hard Causal Constraint + Soft Learned Aggregation).}
CamS uses a hard causal mask to enforce fine-to-coarse information flow, as defined by the \textbf{Inter-Scale Order}:
\begin{equation}
    \psi_{ij}^{\text{CamS}} =
    \begin{cases}
    0 & \text{if } i \in \text{Prefix}(j) \\
    -\infty & \text{otherwise.}
    \end{cases}
\end{equation}
\begin{itemize}
    \item \textbf{Bottom-up Composition:} The Inter-Scale Order ensures that coarse motifs attend to their fine-grained constituents.
    \item \textbf{Anisotropic Learning:} Within the permitted prefix ($\psi_{ij}=0$), connectivity is content-adaptive via $\mathbf{q}_i^\top\mathbf{k}_j$. This allows the model to dynamically select relevant fine-scale neighborhood details (e.g., focusing on a specific pharmacophore) rather than uniformly mixing neighbors, yielding a learned, anisotropic aggregation.
\end{itemize}

\FloatBarrier
\clearpage

% =========================================================
% Appendix C: Training Details & Benchmarks
% =========================================================
\section{Training Details and Benchmark Descriptions}
\label{app:training_impl}

\textbf{Overview.} This section supplements the experimental setup in Sec.~\ref{sec:experiment}. It provides:
(1) Implementation details for the shared training framework (Sec.~\ref{app:training_details});
(2) Complete hyperparameter configurations for pre-training (Table~\ref{tab:pretrain_hparams}) and fine-tuning (Tables~\ref{tab:finetune_fixed} and \ref{tab:finetune_grid});
(3) Detailed descriptions of the downstream tasks in MoleculeNet and MoleculeACE (Sec.~\ref{app:benchmark_details}).

\subsection{Pre-training and Fine-tuning Implementation}
\label{app:training_details}

\paragraph{Framework.}
All experiments are implemented in PyTorch using HuggingFace Transformers.
Pre-training utilizes the \texttt{Trainer} API with DeepSpeed ZeRO-Stage 2 for multi-GPU data parallelism. Fine-tuning runs as single-GPU jobs using strategy-specific trainer subclasses.

\paragraph{Pre-training Configuration.}
We train a 16-layer LLaMA-style decoder initialized from scratch (random initialization).
The training corpus (Enamine675M augmented to $\sim$3.4B views) is pre-tokenized and stored as memory-mapped datasets.
We use FP16 mixed precision, a cosine learning rate schedule with warmup, and periodic evaluation.
The specific hyperparameters are listed in Table~\ref{tab:pretrain_hparams}.

\paragraph{Fine-tuning Protocol.}
We fine-tune the pre-trained backbone using the Dual-Path Strategy (Section~\ref{sec:method_cams_llama}), injecting fingerprints (specifically \textbf{RDKit topological fingerprints}, default parameters, 2048 bits) via a linear projection layer (\texttt{fp\_dim}=2048).
We employ two dropout schemes during fine-tuning (searched via grid search):
(1) \textbf{Head-only Dropout}: Applied only to the task-specific prediction head.
(2) \textbf{Backbone-synced Dropout}: Applied to the head and overriding all backbone dropout rates (attention and hidden states).
Metrics are task-dependent: AUROC for classification (higher is better) and RMSE for regression (lower is better).

\begin{table}[H]
\caption{Pre-training hyperparameters on Enamine675M.}
\label{tab:pretrain_hparams}
\vskip 0.1in
\begin{center}
\begin{small}
\begin{tabular}{ll}
\toprule
\textbf{Hyperparameter} & \textbf{Value} \\
\midrule
Backbone Architecture & LLaMA Decoder (16 layers, 720 hidden size, 8 heads, 2880 MLP dim) \\
Context Length & 4096 tokens \\
Initialization & Random (from local config) \\
Tokenizer & CamS-Tokenizer (Vocab size $\approx$ 67K) \\
Data Augmentation & 5 views per molecule (4 single-scale + 1 multi-scale) \\
\midrule
Parallelism & DeepSpeed ZeRO Stage 2 (8 GPUs) \\
Precision & FP16 Mixed Precision \\
Micro-batch Size & 256 per GPU \\
Gradient Accumulation & 2 steps \\
Global Batch Size & 4096 sequences ($256 \times 8 \times 2$) \\
Optimizer & AdamW ($\beta_1=0.9, \beta_2=0.999$) \\
Peak Learning Rate & $1 \times 10^{-4}$ \\
Weight Decay & $1 \times 10^{-2}$ \\
LR Schedule & Cosine with 5000 warmup steps \\
Gradient Clipping & 1.0 \\
Training Duration & \textbf{Two-stage training}: Stage 1 (1 epoch) + Stage 2 (0.1 epoch with adjusted LR). Total samples seen $\approx$ 3.7B (where 1 epoch = 3.4B augmented views). \\
\bottomrule
\end{tabular}
\end{small}
\end{center}
\vskip -0.1in
\end{table}

\begin{table}[H]
\caption{Fine-tuning setup for downstream tasks.}
\label{tab:finetune_fixed}
\vskip 0.1in
\begin{center}
\begin{small}
\begin{tabular}{lll}
\toprule
Setting & MoleculeNet & MoleculeACE \\
\midrule
Task Type & Classification / Regression & Regression \\
Split Strategy & Scaffold Split (KPGT protocol) & Scaffold Split (Benchmark default) \\
Epochs & 80 & 50 \\
Batch Size & 32 (per GPU) & 32 (per GPU) \\
Optimizer & AdamW & AdamW \\
LR Schedule & Linear decay (warmup ratio 0.1) & Linear decay (warmup ratio 0.1) \\
Metric & ROC-AUC (Cls) / RMSE (Reg) & RMSE \\
\bottomrule
\end{tabular}
\end{small}
\end{center}
\vskip -0.1in
\end{table}

\begin{table}[H]
\caption{Hyperparameter search space for fine-tuning.}
\label{tab:finetune_grid}
\vskip 0.1in
\begin{center}
\begin{small}
\begin{tabular}{ll}
\toprule
Hyperparameter & Search Grid \\
\midrule
Learning Rate & $\{1\!\times\!10^{-6}, 3\!\times\!10^{-6}, 1\!\times\!10^{-5}, 3\!\times\!10^{-5}\}$ \\
Dropout Rate & $\{0.0, 0.05, 0.1, 0.2\}$ \\
Weight Decay & $\{0.0, 1\!\times\!10^{-6}, 1\!\times\!10^{-4}\}$ \\
Fine-tune Strategy & \texttt{Standard}, \texttt{L2SP}, \texttt{LLRD}, \texttt{FLAG}, \texttt{Reinit} \\
\bottomrule
\end{tabular}
\end{small}
\end{center}
\vskip -0.1in
\end{table}

\subsection{Benchmark Task Descriptions}
\label{app:benchmark_details}

We evaluate CamS-LLaMA on two complementary benchmarks to assess both general property prediction and structural sensitivity.

\paragraph{MoleculeNet (General Properties) \cite{wu2018moleculenet}.}
This benchmark covers a diverse range of molecular properties. We select 11 datasets that cover physiology, biophysics, and physical chemistry. Consistent with our results table, the tasks are:

\begin{itemize}
    \item \textbf{Classification Tasks (8):}
    \begin{itemize}
        \item \textbf{BACE}: Inhibition of $\beta$-secretase (a key Alzheimer's therapeutic target).
        \item \textbf{BBBP}: Blood-brain barrier penetration (permeability).
        \item \textbf{ClinTox}: Clinical toxicity (distinguishing FDA-approved drugs from toxic compounds).
        \item \textbf{Estrogen}: Estrogen receptor ($\alpha$, $\beta$) binding activity (endocrine disruption potential).
        \item \textbf{Metstab}: Metabolic stability (half-life duration in liver microsomes).
        \item \textbf{SIDER}: Adverse drug reactions (side effects) of marketed medicines.
        \item \textbf{ToxCast}: High-throughput toxicology screening data.
        \item \textbf{Tox21}: Toxicity testing across 12 biological targets (nuclear receptors/stress pathways).
    \end{itemize}
    \item \textbf{Regression Tasks (3):}
    \begin{itemize}
        \item \textbf{ESOL}: Water solubility (log solubility in mols per litre).
        \item \textbf{FreeSolv}: Hydration free energy (experimental vs calculated).
        \item \textbf{Lipo}: Lipophilicity (octanol/water distribution coefficient, logD).
    \end{itemize}
\end{itemize}
All datasets follow the \textbf{scaffold splitting} strategy (as used in KPGT) to strictly evaluate the model's generalization capability to chemically distinct structures.

\paragraph{MoleculeACE (Activity Cliffs) \cite{van2022exposing}.}
MoleculeACE is designed to stress-test models on Activity Cliffs—pairs of molecules with high structural similarity but large differences in potency.
It consists of 30 datasets derived from ChEMBL, each targeting a specific biological protein target (e.g., \textit{CHEMBL204}, \textit{CHEMBL240}).
\begin{itemize}
    \item \textbf{Challenge:} Unlike MoleculeNet, which often rewards global scaffold recognition, MoleculeACE requires the model to identify fine-grained structural edits (e.g., methylation, halogenation) that trigger drastic activity shifts.
    \item \textbf{Metric:} Performance is measured by RMSE on the test set (scaffold split). Lower RMSE indicates better ability to capture the non-smooth structure-activity landscape.
\end{itemize}

\paragraph{Evaluation Protocols and Statistical Reporting.}
We strictly adhere to the standard evaluation protocols specific to each benchmark, leading to different reporting formats for statistical significance:
\textbf{On MoleculeNet (3 Random Seeds),} following the protocol established by KPGT \cite{li2023knowledge}, we employ \textbf{scaffold splitting} with an 8:1:1 ratio. To ensure robust estimation, we repeat all experiments over \textbf{3 independent random seeds}. Consequently, results in Table~\ref{tab:MoleculeNet_final} and Table~\ref{tab:app_MoleculeNet} are reported as \textbf{Mean$_{\text{(SD)}}$}, capturing the variance arising from data splitting.
\textbf{On MoleculeACE,} this benchmark provides a \textbf{pre-defined, deterministic scaffold split} for each task to rigorously standardize the evaluation of specific activity-cliff pairs \cite{van2022exposing}. We evaluate the model exactly once on this fixed official test set. Since the evaluation involves no random resampling, \textbf{Standard Deviation is not applicable}, and we report the exact performance values (Table~\ref{tab:moleculeace_final} and Table~\ref{tab:app_moleculeace}).

\FloatBarrier
\clearpage

% =========================================================
% Appendix D: Comparison with Large-scale SMILES FMs
% =========================================================
\section{Comparison with Large-scale SMILES FMs}
\label{app:naturelm}
To explicitly address the comparison with standard SMILES-based NTP foundation models, we include the recently reported results of \textbf{NatureLM} \cite{xia2025nature}. 
NatureLM is a SMILES-based FM trained on 3.4 billion molecules with model sizes ranging from 1B to 8$\times$7B parameters.
As shown in Table~\ref{tab:naturelm_comparison}, despite being orders of magnitude larger in both parameter count ($>$100$\times$) and training data ($>$5$\times$) compared to CamS-LLaMA ($\sim$0.1B, 675M data), NatureLM significantly underperforms on discriminative tasks. 
This empirical gap serves as a strong validation that pure SMILES-based NTP, even at extreme scales, struggles to capture the structural features required for property prediction, necessitating the topological enhancement provided by CamS.

\begin{table}[h]
\caption{Comparison with a SOTA SMILES-based Natural Language Science Foundation Models (NatureLM \cite{xia2025nature}). Results for NatureLM are taken directly from the original paper (Table 19). CamS-LLaMA ($\sim$0.1B) significantly outperforms NatureLM (up to 56B parameters) on common benchmarks, demonstrating superior representational efficiency.}
\label{tab:naturelm_comparison}
\begin{center}
\begin{small}
\begin{sc}
\begin{tabular}{lcccc}
\toprule
Model & Params & BBBP & BACE & Tox21 \\
\midrule
NatureLM & 1B & 0.711 & 0.794 & 0.683 \\
NatureLM & 8B & 0.702 & 0.820 & 0.698 \\
NatureLM & 8$\times$7B (MoE) & 0.737 & 0.831 & 0.720 \\
\midrule
\textbf{CamS-LLaMA} & \textbf{0.1B} & \textbf{0.942} & \textbf{0.870} & \textbf{0.827} \\
\bottomrule
\end{tabular}
\end{sc}
\end{small}
\end{center}
\end{table}

% =========================================================
% Appendix E: Extended Discussion on Data Scale, Fairness, and Efficiency
% =========================================================
\section{Extended Discussion on Data Scale, Fairness, and Efficiency}
\label{app:data_scale}

A prominent distinction between CamS-LLaMA and graph-native baselines (e.g., KPGT) is the pre-training corpus size (675M vs.\ 550K). We emphasize that this disparity does not constitute experimental unfairness, but rather demonstrates a critical \textbf{methodological advantage} of our sequence-based framework: \textbf{Scalability}.

Graph-native foundation models typically rely on heavy auxiliary inputs or targets to compensate for the lack of semantic density in raw graphs. This creates distinct scaling barriers:
\begin{itemize}
    \item \textbf{Descriptor Calculation (e.g., KPGT):} KPGT requires computing over 200 RDKit descriptors and fingerprints for \textit{every} training instance to serve as "knowledge" targets. Scaling this dense annotation to the 675M-scale Enamine dataset (let alone augmented views) imposes significant data engineering and storage overheads.
    \item \textbf{3D Conformer Generation (e.g., GEM, Uni-Mol):} Other baselines like GEM explicitly rely on 3D geometric views. Generating conformers (via ETKDG or DFT) allows for richer physics but is computationally prohibitive at the billion-scale.
\end{itemize}
These pre-processing costs limit such methods to smaller datasets (e.g., $\sim$500K) by design necessity.

In contrast, CamS transforms molecular graphs into causal sequences via purely logical operations (BPE merge and BFS traversal). These operations involve only string manipulation and standard graph traversal, which are CPU-efficient and easily parallelizable. CamS operates directly on pure molecular topology without requiring external descriptor injection or 3D optimization during pre-training.
This efficiency removes the data-scaling bottleneck, allowing our method to leverage billion-scale augmented datasets as a standard feature of its training pipeline.

\paragraph{Diversity over Repetition (High-Coverage Training).} Our model was trained for $\sim$1 epoch on the augmented Enamine dataset (5 views per molecule).
While baselines typically train on a small dataset (550K) for many epochs (High Repetition), CamS processes a vast number of unique structures (675M) with limited repetition per structure (High Diversity).
The performance superiority of CamS, therefore, stems from its ability to efficiently access and learn from the \textbf{breadth of the chemical space}. This "Diversity over Repetition" regime is unlocked specifically by the efficient nature of the CamS representation, which would be computationally infeasible for descriptor-heavy graph baselines.

% =========================================================
% Appendix F: Detailed Main Results (kept content)
% =========================================================
\section{Detailed Experimental Results}
\label{app:detailed_results}
\textbf{Overview.} This section corresponds to the benchmark results summarized in Sec.~\ref{sec:experiment}. It supplements the main text with the full detailed tables of (1) MoleculeNet overall performance of all baseline and their rank comparison (extended version of Table~\ref{tab:MoleculeNet_final}; Table~\ref{tab:app_MoleculeNet}), and (2) MoleculeACE results overall performance of all baseline and their rank comparison (extended version of Table~\ref{tab:moleculeace_final}; Table~\ref{tab:app_moleculeace}).

\begin{table}[H]
\caption{Full Performance comparison on MoleculeNet benchmark. (Extended version of Table~\ref{tab:MoleculeNet_final}). Sup variants:  $_{\text{sup}}$ denotes adding an additional supervised graph-level bioactivity pre-training stage on top of the corresponding self-supervised objective (same protocol as used in the baseline setting). Baseline Reference: Infomax~\cite{velivckovic2018deep}; Edgepred~\cite{hamilton2017inductive}; Attribute Masking (Masking)~\cite{hu2020strategies}; Context Prediction (Contextpred)~\cite{hu2020strategies}; GraphLoG~\cite{xu2021self}; GraphCL~\cite{you2020graph}; JOAO~\cite{you2021graph}; GROVER~\cite{rong2020self}; 3DInfomax~\cite{stark20223d}; GraphMVP~\cite{liu2016pyridazinone}; ImageMol~\cite{zeng2022accurate}; MolFormer~\cite{wu2023molformer}; GEM~\cite{fang2022geometry}; GraphMAE~\cite{hou2022graphmae}; MoleBert~\cite{xia2023mole}; KPGT~\cite{li2023knowledge}}
\label{tab:app_MoleculeNet}
\vskip 0.15in
\begin{center}
\begin{small}
\setlength{\tabcolsep}{4pt}
\resizebox{\textwidth}{!}{
\begin{tabular}{lcclcc}
\toprule
\multicolumn{3}{c}{Classification Tasks} &
\multicolumn{3}{c}{Regression Tasks} \\
\cmidrule(lr){1-3}\cmidrule(lr){4-6}
Method & AVG-AUROC $\uparrow$ & Rank & Method & AVG-RMSE $\downarrow$ & Rank \\
\midrule
CamS-LLaMA & 0.845 & 1 & CamS-LLaMA & 1.172 & 1 \\
KPGT & 0.843 & 2 & KPGT & 1.175 & 2 \\
CamS-LLaMA (w/o FP) & 0.838 & 3 & CamS-LLaMA (w/o FP) & 1.195 & 3 \\
CamS-LLaMA (Vocab-1K Only) & 0.833 & 4 & CamS-LLaMA (Vocab-1K Only) & 1.215 & 4 \\
GEM & 0.825 & 5 & MolFormer & 1.272 & 5 \\
CamS-LLaMA (Vocab-67K Only) & 0.818 & 6 & GEM & 1.285 & 6 \\
GROVER & 0.818 & 7 & CamS-LLaMA (Vocab-67K Only) & 1.329 & 7 \\
Contextpred$_{Sup}$ & 0.807 & 8 & GROVER & 1.332 & 8 \\
3DInfomax & 0.804 & 9 & 3DInfomax & 1.400 & 9 \\
ImageMol & 0.802 & 10 & Contextpred$_{Sup}$ & 1.414 & 10 \\
MoleBERT & 0.802 & 11 & ImageMol & 1.501 & 11 \\
Masking$_{Sup}$ & 0.799 & 12 & MoleBERT & 1.559 & 12 \\
Edgepred$_{Sup}$ & 0.795 & 13 & Contextpred & 1.563 & 13 \\
Infomax$_{Sup}$ & 0.794 & 14 & Edgepred$_{Sup}$ & 1.576 & 14 \\
JOAO & 0.790 & 15 & Masking$_{Sup}$ & 1.620 & 15 \\
Contextpred & 0.790 & 16 & GraphMVP & 1.647 & 16 \\
GraphMAE & 0.788 & 17 & Infomax$_{Sup}$ & 1.661 & 17 \\
Edgepred & 0.785 & 18 & GraphLoG & 1.663 & 18 \\
GraphCL & 0.774 & 19 & Edgepred & 1.674 & 19 \\
Infomax & 0.773 & 20 & Masking & 1.697 & 20 \\
Masking & 0.773 & 21 & GraphMAE & 1.716 & 21 \\
GraphLoG & 0.769 & 22 & Infomax & 1.770 & 22 \\
GraphMVP & 0.769 & 23 & GraphCL & 1.822 & 23 \\
MolFormer & 0.744 & 24 & JOAO & 1.960 & 24 \\
\bottomrule
\end{tabular}
}
\end{small}
\end{center}
\vskip -0.1in
\end{table}

\begin{table}[t]
\caption{Full Detailed results on 30 MoleculeACE activity-cliff tasks (Regression, RMSE $\downarrow$). (Extended version of Table~\ref{tab:moleculeace_final}). Sup variants:  $_{\text{sup}}$ denotes adding an additional supervised graph-level bioactivity pre-training stage on top of the corresponding self-supervised objective (same protocol as used in the baseline setting). Descriptors: ECFP~\cite{rogers2010extended}; MACCS~\cite{durant2002reoptimization}; PHYSCHEM~\cite{walters2002prediction}; WHIM~\cite{kubinyi19933d}. Baseline Reference: support vector machines (SVM)~\cite{cristianini2002support}; random forest (RF)~\cite{breiman1996bagging}; gradient boosting machine (GBM)~\cite{friedman2001greedy}; k-nearest neighbor (KNN)~\cite{fix1985discriminatory}; message passing neural network (MPNN)~\cite{gilmer2017neural}; graph attention network (GAT)~\cite{velivckovic2017graph}; graph convolutional network (GCN)~\cite{kipf2016semi}; AFP~\cite{xiong2019pushing}; Convolutional Neural NetworkLong Short-Term Memory (CNN, specifically Maxsmi) ~\cite{kimber2021maxsmi}; Long Short-Term Memory (LSTM, specifically CLM) ~\cite{moret2022perplexity}; Transformer (specifically Chemberta)~\cite{chithrananda2020chemberta}; KPGT~\cite{li2023knowledge}}
\label{tab:app_moleculeace}
\vskip 0.15in
\begin{center}
\begin{scriptsize}
\setlength{\tabcolsep}{3.2pt}
\resizebox{\textwidth}{!}{
\begin{tabular}{lcclcc}
\toprule
\multicolumn{3}{c}{MoleculeACE AVG-RMSE (Ranks 1--24)} & \multicolumn{3}{c}{MoleculeACE AVG-RMSE (Ranks 25--48)} \\
\cmidrule(lr){1-3}\cmidrule(lr){4-6}
Method & AVG-RMSE $\downarrow$ & Rank & Method & AVG-RMSE $\downarrow$ & Rank \\
\midrule
CamS-LLaMA & 0.624 & 1 & EdgePred$_{Sup}$ & 0.764 & 25 \\
KPGT & 0.633 & 2 & Contextpred$_{Sup}$ & 0.764 & 26 \\
CamS-LLaMA (Vocab-1K Only) & 0.641 & 3 & GraphMVP & 0.768 & 27 \\
CamS-LLaMA (Vocab-67K Only) & 0.649 & 4 & Infomax & 0.774 & 28 \\
CamS-LLaMA (w/o FP) & 0.650 & 5 & Edgepred & 0.774 & 29 \\
SVM$_{ECFP}$ & 0.675 & 6 & ImageMol & 0.797 & 30 \\
GROVER & 0.680 & 7 & 3DInfomax & 0.804 & 31 \\
GBM$_{ECFP}$ & 0.701 & 8 & GraphLoG & 0.807 & 32 \\
RF$_{ECFP}$ & 0.705 & 9 & KNN$_{MACCS}$ & 0.818 & 33 \\
MolFormer & 0.740 & 10 & GEM & 0.821 & 34 \\
KNN$_{ECFP}$ & 0.741 & 11 & Transformer & 0.868 & 35 \\
MLP$_{ECFP}$ & 0.742 & 12 & RF$_{PHYSCHEM}$ & 0.890 & 36 \\
GBM$_{MACCS}$ & 0.742 & 13 & GBM$_{PHYSCHEM}$ & 0.901 & 37 \\
LSTM & 0.744 & 14 & KNN$_{PHYSCHEM}$ & 0.923 & 38 \\
Contextpred & 0.747 & 15 & SVM$_{PHYSCHEM}$ & 0.935 & 39 \\
RF$_{MACCS}$ & 0.753 & 16 & CNN & 0.937 & 40 \\
GraphMAE & 0.753 & 17 & MPNN & 0.959 & 41 \\
SVM$_{MACCS}$ & 0.754 & 18 & AFP & 0.970 & 42 \\
Masking$_{Sup}$ & 0.755 & 19 & RF$_{WHIM}$ & 0.977 & 43 \\
MoleBERT & 0.755 & 20 & GBM$_{WHIM}$ & 0.992 & 44 \\
Infomax$_{Sup}$ & 0.756 & 21 & SVM$_{WHIM}$ & 1.003 & 45 \\
JOAO & 0.757 & 22 & GCN & 1.010 & 46 \\
Masking & 0.758 & 23 & KNN$_{WHIM}$ & 1.020 & 47 \\
GraphCL & 0.760 & 24 & GAT & 1.049 & 48 \\
\bottomrule
\end{tabular}
}
\end{scriptsize}
\end{center}
\vskip -0.1in
\end{table}

\clearpage

% =========================================================
% Appendix G: Interpretability Details (Main-text ref exists!)
% =========================================================
\section{Interpretability Details for Activity-Cliff Attention Analysis}
\label{app:interpret}

\textbf{Overview.}
This section provides implementation details for the attention analysis presented in Section~\ref{sec:interpretability}.
It complements the main text by specifying:
(1) the exact algorithm for identifying differential atoms in activity-cliff pairs (Appendix~\ref{app:interpret_pairs}, Algorithm~\ref{alg:ac_shared_diff_fragments});
(2) the mapping from atom-level labels to CamS tokens across different \textbf{Motif Scales} (Appendix~\ref{app:interpret_align});
and (3) the detailed computation of the Rel-DTAP metric (Appendix~\ref{app:interpret_metric}).

\subsection{Activity-cliff Pair Construction and Differential/Shared Atom Identification}
\label{app:interpret_pairs}
% TODO: MoleculeACE 原文的 cliff pair 构造流程；阈值/定义；anchor/partner 是否分别计数（你正文说 separately counted）
% TODO: diff/shared atom-level point 的判定算法（按你实现写清）
\paragraph{Pair construction (MoleculeACE-style).}
For each MoleculeACE sub-task dataset, we start from its per-molecule CSV containing \texttt{smiles}, \texttt{exp\_mean [nM]}, \texttt{cliff\_mol}, and the split label.
We treat molecules with \texttt{cliff\_mol=1} as anchors (by default restricted to the test split), and search partners among molecules with \texttt{cliff\_mol=0} (by default from the same split).
A candidate pair is kept if it satisfies both (1) \textbf{high structural similarity} and (2) \textbf{large potency change}.
Structural similarity is a ``soft-consensus'' rule: we compute (a) full-molecule ECFP Tanimoto, (b) generic Murcko-scaffold ECFP Tanimoto, and (c) SMILES Levenshtein similarity, and accept the pair if \textbf{any} of them is $\ge\tau_{\mathrm{sim}}$ (default $0.9$).
Potency change is measured by the fold change on linear \texttt{exp\_mean [nM]} values:
$
\mathrm{FC}=\frac{\max(y_a,y_p)}{\max(\min(y_a,y_p),\epsilon)}
$
(with $\epsilon{=}10^{-12}$), and we keep the pair if $\mathrm{FC}\ge\tau_{\mathrm{fold}}$ (default $10$).
Each selected pair yields one record $(\text{anchor},\text{partner})$; for downstream attention statistics we count the two molecules (anchor and partner) separately, as stated in Sec.~\ref{sec:interpretability}.

\paragraph{Differential vs.\ shared atoms.}
Given an (anchor, partner) pair, we identify shared atoms via an RDKit maximum common substructure (MCS) query with ring constraints and chemistry-aware matching (elements compared by type; bonds compared by order; \texttt{completeRingsOnly} and \texttt{ringMatchesRingOnly} enabled).
Let $\mathcal{A}$ and $\mathcal{B}$ be the atom-index sets of the two molecules and let $\mathcal{M}\subseteq\mathcal{A}$ and $\mathcal{M}'\subseteq\mathcal{B}$ be the matched atom indices returned by the first substructure match of the MCS query.
We define \textbf{differential atoms} as the complement sets $\mathcal{A}_{\Delta}=\mathcal{A}\setminus\mathcal{M}$ and $\mathcal{B}_{\Delta}=\mathcal{B}\setminus\mathcal{M}'$, and \textbf{shared atoms} as $\mathcal{M}$ and $\mathcal{M}'$.
If MCS fails (or a SMILES cannot be parsed), we conservatively treat all atoms as differential.

\paragraph{Atom correspondence (optional for visualization).}
For visualization/debugging we also record the ordered atom correspondence induced by the MCS query, i.e., a list of paired indices $\{(i,j)\}$ obtained by aligning the two match tuples in the MCS SMARTS query order.
This mapping is not required for computing Rel-DTAP, but enables atom-level highlighting across the two molecules.

\subsection{Mapping Atom-level Diff/Shared Labels to Tokens in Each Scale Region}
\label{app:interpret_align}

\paragraph{Token-to-atom alignment.}
We re-encode each molecule with \texttt{MultiGraphBPETokenizerExplain}, which returns, for every token position $t$, an atom-index set $S_t$ indicating which RDKit atom indices are covered by that token (special tokens such as \texttt{[BOS]}, \texttt{[EOS]}, and \texttt{[CONCAT]} have empty sets).
Because GraphBPE encoding inserts auxiliary connector atoms (\texttt{*}) with indices outside the original RDKit atom range, we remove such indices by clipping $S_t$ to $\{0,\dots,N{-}1\}$ where $N$ is the number of atoms in the original molecule.

\paragraph{From atom-level labels to token-level labels.}
Let $\mathcal{A}_\Delta$ be the differential-atom set for this molecule from Appendix~\ref{app:interpret_pairs}.
We assign a token-level differential indicator by an \textbf{any-diff} rule:
a token at position $t$ is marked as differential iff $S_t \cap \mathcal{A}_\Delta \neq \varnothing$; otherwise it is marked as shared/non-differential.
Thus a token covering multiple atoms is counted as differential if it contains at least one differential atom.

\paragraph{Scale-region boundaries in the concatenated sequence.}
For multi-scale CamS sequences, scale regions are delimited by the \texttt{[CONCAT]} separators.
Concretely, given the full token-id list (including \texttt{[BOS]}, \texttt{[EOS]}, and \texttt{[CONCAT]}), we split it into consecutive spans between \texttt{[CONCAT]} tokens, and exclude \texttt{[BOS]/[EOS]/[CONCAT]} themselves from the per-region statistics.
This yields the four scale regions (e.g., 1K/7K/27K/67K in the paper) used to report scale-wise attention preferences.

\subsection{Attention Extraction and Metric Computation}
\label{app:interpret_metric}

\paragraph{Attention extraction protocol.}
For each molecule, we run a forward pass with \texttt{output\_attentions=True} and extract the \textbf{last-layer} attention tensor.
We average over heads to obtain a single attention matrix $\bar{\mathbf{A}}\in\mathbb{R}^{S\times S}$, where $S$ is the sequence length.
We use the attention distribution of the final token (the last position in the sequence) as a saliency-like weighting over the prefix tokens, i.e., the row vector $\bar{\mathbf{A}}_{S,:}$.
We compute this distribution in two modes:
(1) \textbf{without fingerprint input} by running the model without the prepended fingerprint token;
(2) \textbf{with fingerprint input} by prepending the fingerprint embedding (sequence length becomes $S{+}1$), extracting $\bar{\mathbf{A}}_{S{+}1,:}$, and then taking only the sub-vector over molecular tokens (excluding the fingerprint position) for diff/shared statistics.

\paragraph{MDTA/MSTA within a scale region.}
Within a given scale region $s$ (defined by \texttt{[CONCAT]} boundaries), let $p_i$ denote the (renormalized) attention weight on token $i$ in that region, and let $d_i\in\{0,1\}$ be its differential indicator from Appendix~\ref{app:interpret_align}.
We define the mean differential-token attention and mean shared-token attention as
\[
\mathrm{MDTA}_s=\frac{\sum_i p_i d_i}{\sum_i d_i},\qquad
\mathrm{MSTA}_s=\frac{\sum_i p_i (1-d_i)}{\sum_i (1-d_i)},
\]
with the convention that the corresponding mean is set to $0$ if the denominator is $0$ (no tokens of that type in the region).

\paragraph{Rel-DTAP computation and aggregation.}
For each molecule (anchor and partner counted separately), we compute
\[
\mathrm{Rel\text{-}DTAP}_s=\frac{\mathrm{MDTA}_s-\mathrm{MSTA}_s}{\mathrm{MSTA}_s+\epsilon}\times 100,
\]
using $\epsilon=10^{-12}$ for numerical stability.
We report the final Rel-DTAP by averaging this quantity over all molecules from activity-cliff pairs, both for each scale region $s$ and for the full concatenated sequence (all regions combined).

% Shared/diff fragment (token) labeling for an activity-cliff pair
\begin{algorithm}[H]
\caption{Shared/differential fragment labeling for an activity-cliff pair}
\label{alg:ac_shared_diff_fragments}
\begin{algorithmic}[1]
\STATE \textbf{Input:} SMILES pair $(s_a,s_b)$; RDKit MCS; tokenizer-explain encoder $\iota$ returning per-token atom sets
\STATE \textbf{Output:} shared/diff atom sets $(M_a,\Delta_a)$ and $(M_b,\Delta_b)$; token-level diff masks $\mathbf{d}^{(a)},\mathbf{d}^{(b)}$; optional atom map $\mathcal{P}$

\STATE $m_a\leftarrow \mathrm{MolFromSmiles}(s_a)$; $m_b\leftarrow \mathrm{MolFromSmiles}(s_b)$
\STATE $N_a\leftarrow m_a.\mathrm{GetNumAtoms}()$; $N_b\leftarrow m_b.\mathrm{GetNumAtoms}()$

\STATE $smarts \leftarrow \mathrm{FindMCS}([m_a,m_b];$
\item[] \hspace{1.5em} \texttt{completeRingsOnly}, \texttt{ringMatchesRingOnly},
\item[] \hspace{1.5em} \texttt{CompareElements}, \texttt{CompareOrder}).\texttt{smartsString}
\IF{$smarts$ is empty}
    \STATE $M_a\leftarrow\varnothing$; $M_b\leftarrow\varnothing$
\ELSE
    \STATE $q\leftarrow \mathrm{MolFromSmarts}(smarts)$
    \STATE $\mathbf{t}_a\leftarrow m_a.\mathrm{GetSubstructMatch}(q)$; $\mathbf{t}_b\leftarrow m_b.\mathrm{GetSubstructMatch}(q)$ \hfill (take the first match)
    \STATE $M_a\leftarrow \mathrm{set}(\mathbf{t}_a)$; $M_b\leftarrow \mathrm{set}(\mathbf{t}_b)$
    \STATE $\mathcal{P}\leftarrow \{(\mathbf{t}_a[i],\mathbf{t}_b[i])\}_{i=1}^{|\mathbf{t}_a|}$ \hfill (ordered atom correspondence)
\ENDIF
\STATE $\Delta_a\leftarrow \{0,\dots,N_a{-}1\}\setminus M_a$; $\Delta_b\leftarrow \{0,\dots,N_b{-}1\}\setminus M_b$

\FOR{side $m\in\{a,b\}$}
    \STATE Encode $s_m$ with $\iota$ to get tokens $(x^{(m)}_1,\dots,x^{(m)}_{L_m})$ and atom sets $\{S^{(m)}_t\}_{t=1}^{L_m}$
    \STATE Clip connector atoms: $S^{(m)}_t \leftarrow S^{(m)}_t \cap \{0,\dots,N_m{-}1\}$ for all $t$
    \FOR{$t=1$ \textbf{to} $L_m$}
        \IF{$x^{(m)}_t$ is \texttt{[BOS]} or \texttt{[EOS]} or \texttt{[CONCAT]}}
            \STATE $d^{(m)}_t \leftarrow 0$
        \ELSE
            \IF{$S^{(m)}_t \cap \Delta_m \neq \varnothing$}
                \STATE $d^{(m)}_t \leftarrow 1$ \hfill (diff fragment token; any-diff rule)
            \ELSE
                \STATE $d^{(m)}_t \leftarrow 0$ \hfill (shared fragment token)
            \ENDIF
        \ENDIF
    \ENDFOR
\ENDFOR
\end{algorithmic}
\end{algorithm}

\subsection{Case-Study Pair Selection Protocol}
\label{app:case_selection}

To ensure the representativeness and reproducibility of the qualitative analysis in Section~\ref{sec:interpretability}, we employ a rigorous, log-driven selection pipeline rather than manual cherry-picking.
We build case studies from the test-set activity-cliff pairs identified in Appendix~\ref{app:interpret_pairs}.
We implement three complementary selection modes:

\begin{enumerate}
    \item \textbf{Mode A: ``Similar-structure'' cases (Targeting Minimal Edits).}
    This mode isolates pairs that differ by only a few atoms but yield a large potency change.
    \begin{itemize}
        \item \textit{Atom-count prefilter}: We require the atom count difference $\Delta N \le 3$ and pair size $N_{\max} \le 100$.
        \item \textit{MCS-based scoring}: We compute the MCS and define the edit size $d_{\max}$ as the number of non-MCS atoms. We prioritize pairs with $d_{\max} \le 5$.
        \item \textit{Ranking}: Pairs are ranked by increasing edit size (more similar first) and then by decreasing fold change.
    \end{itemize}

    \item \textbf{Mode B: ``Largest fold-change'' cases (Targeting Extreme Cliffs).}
    This mode targets the most extreme potency shifts regardless of edit size. We sort eligible pairs by potency fold-change in descending order and select top candidates.

    \item \textbf{Mode C: ``Relatively larger molecules'' cases (Targeting Complexity).}
    This mode ensures coverage of complex molecules. We filter for pairs with $N_{\max} \ge 40$ and small $\Delta N$, selecting those with the largest molecule sizes.
\end{enumerate}

\paragraph{Justification of Token-Level Labeling (Holistic Chemical Semantics).}
A potential concern is whether the "any-diff" rule (marking a coarse motif as differential if it contains \textit{any} modified atom) biases metrics toward larger tokens.
We argue this design is grounded in \textbf{holistic chemical semantics}. In medicinal chemistry, modifying a single atom within a ring system or functional group (e.g., H $\to$ F on a benzene ring) fundamentally alters the electronic and steric properties of the \textit{entire substructure}.
In CamS, such a modification results in a completely distinct Token ID for the coarse motif. Therefore, high attention to this "diff-containing" coarse token reflects a valid recognition of the \textbf{macro-semantic shift} of the functional group as a whole, rather than a metric artifact.

\subsection{Note on Baselines.} 
The Rel-DTAP metric is specifically designed to evaluate the multi-granular attention allocation inherent to the CamS hierarchy (mapping attention to specific $1\mathrm{K}$ vs.\ $67\mathrm{K}$ token regions). Since standard graph baselines (e.g., KPGT, GROVER) operate on fixed input granularities (atoms or triplets) and lack this explicit hierarchical tokenization, this metric is not directly applicable to them. Thus, our analysis focuses on the intrinsic mechanism of CamS.

\subsection{Additional Case Studies}
\label{app:more_cases}

We provide additional visualizations for pairs selected via Mode A (Minimal Edits) and Mode B (Extreme Cliffs) to further substantiate the findings.

\begin{figure}[h]
    \centering
    \includegraphics[width=\textwidth]{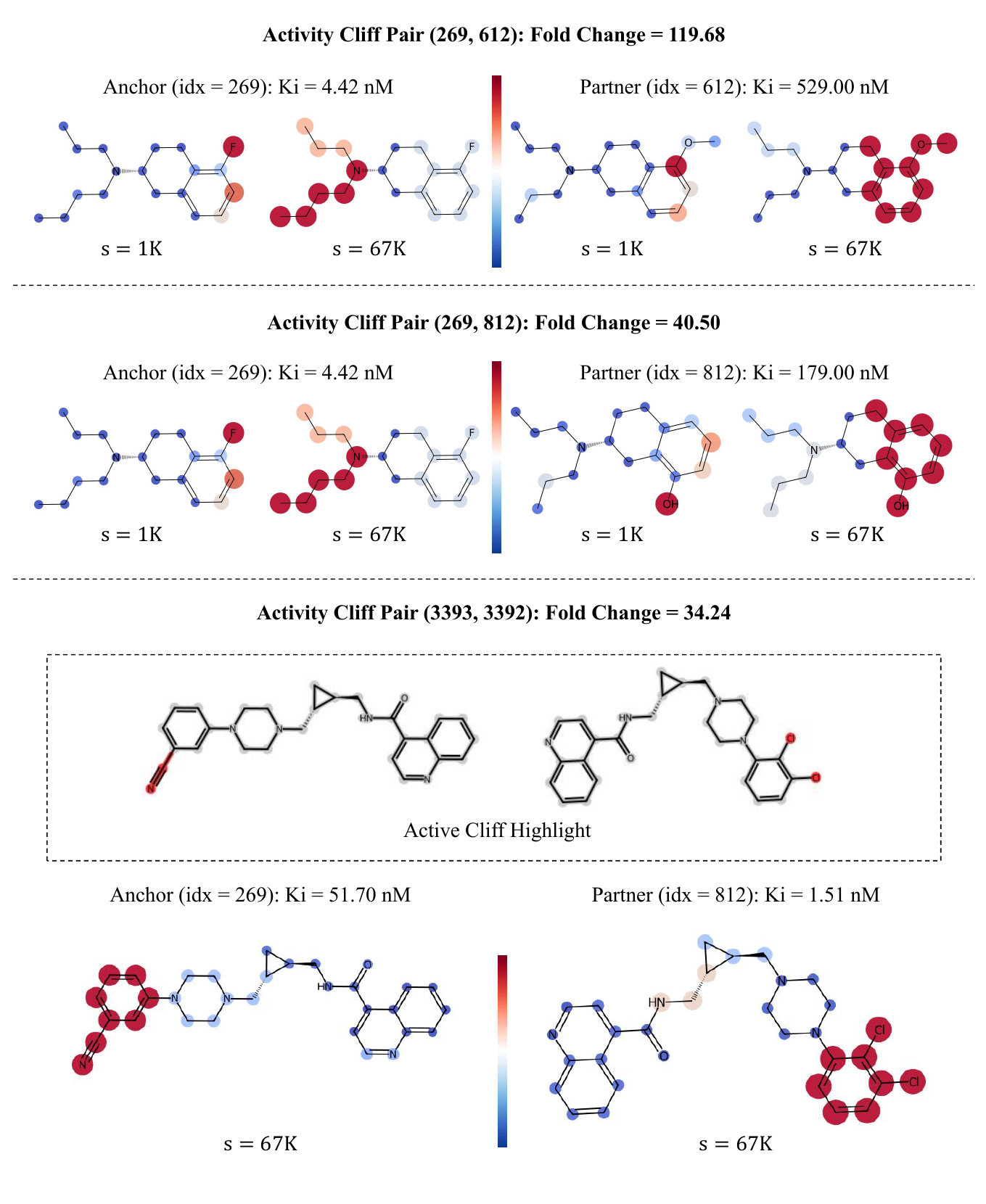}
    \caption{\textbf{Additional Case Study 1 (Minimal Edit).} Attention heatmap for pairs (from CHEMBL234$_{Ki}$) with minimal-atom substitution causing activity cliffs.}
    \label{fig:app_case_1}
\end{figure}

\begin{figure}[h]
    \centering
    \includegraphics[width=\textwidth]{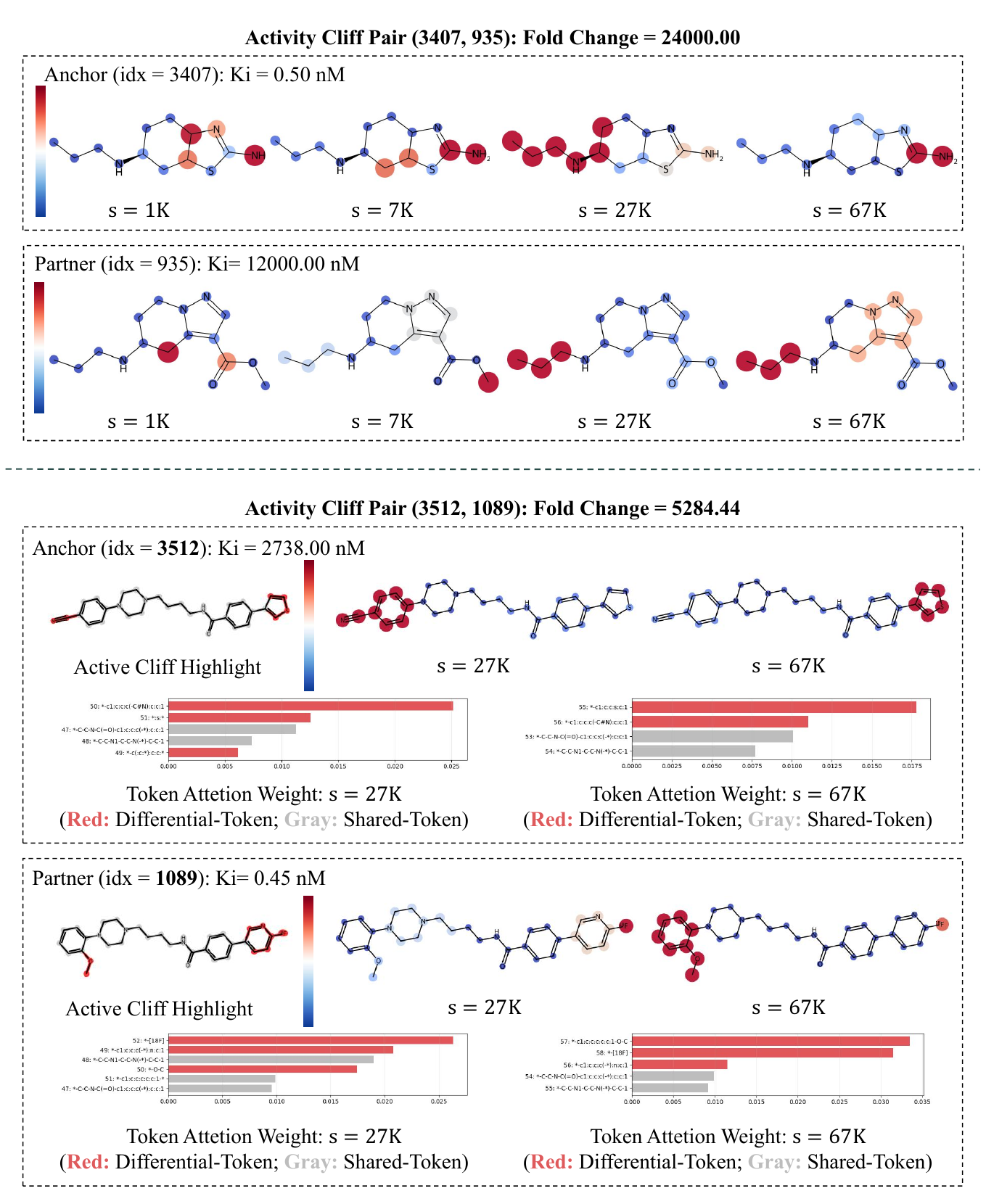}
    \caption{\textbf{Additional Case Study 2 (Extreme Cliff).} Attention heatmap for pairs with a scaffold-hopping modification yielding a $>4000\times$ shift.}
    \label{fig:app_case_2}
\end{figure}

\clearpage

% =========================================================
% Appendix H: Ablation Details 
% =========================================================
\section{Detailed Ablation Analysis}
\label{app:ablation_details}

In this section, we provide the complete breakdown of the ablation study.

\textbf{Budget Matching Protocol.} To ensure a fair comparison, all single-scale variants were trained with a computational budget strictly matched to the full model. Specifically, we controlled the total number of parameter update steps (and thus the total number of training samples seen) to be identical across all settings, eliminating training duration as a confounding factor.

We compare the following variants across all tasks:
\begin{itemize}
    \item \textbf{CamS (Full)}: The proposed multi-scale model with fingerprint injection.
    \item \textbf{w/o FP}: The pure multi-scale sequence model without fingerprint injection.
    \item \textbf{1K Only}: Single-scale model using fine-grained motifs.
    \item \textbf{67K Only}: Single-scale model using coarse-grained motifs.
\end{itemize}

\textbf{1. Impact of Multi-Scale Concatenation.}
As shown in Table~\ref{tab:app_ablation_molnet}, the \textbf{CamS (Full)} model consistently outperforms the single-scale variants (\textbf{1K Only} and \textbf{67K Only}). 
Notably, the \textbf{67K Only} variant suffers significant performance degradation in regression tasks (AVG RMSE 1.329), confirming that excessive compression reduces supervision density.

\textbf{2. Pure Sequence vs. Fingerprint Injection.}
On tasks such as \textbf{Estrogen} and \textbf{ESOL}, the \textbf{w/o FP} variant achieves the best performance (bolded). This indicates that the intrinsic structural features learned by CamS are sometimes superior to explicit fingerprints. However, as shown in Table~\ref{tab:app_ablation_ace}, FP injection is crucial for stability on activity-cliff tasks.

% ==========================================
% Table 11: MoleculeNet Full Ablation with SD
% Added Standard Deviations from raw data
% ==========================================
\begin{table}[h]
\caption{Detailed Ablation on MoleculeNet. Values represent Mean$_{(\text{SD})}$ over 3 random seeds. \textbf{Bold} indicates the best performance among CamS variants.}
\label{tab:app_ablation_molnet}
\begin{center}
\begin{small}
\begin{sc}
% Reduced tabcolsep to fit SDs comfortably within linewidth
\setlength{\tabcolsep}{3pt} 
\begin{tabular}{lccccc}
\toprule
Task & KPGT & \textbf{CamS (Full)} & w/o FP & 1K Only & 67K Only \\
\midrule
\multicolumn{6}{c}{\textit{Classification (AUROC $\uparrow$)}} \\
BACE     & 0.855$_{(0.014)}$ & \textbf{0.870}$_{(0.013)}$ & 0.850$_{(0.014)}$ & 0.845$_{(0.030)}$ & 0.837$_{(0.024)}$ \\
BBBP     & 0.908$_{(0.012)}$ & \textbf{0.942}$_{(0.015)}$ & 0.933$_{(0.021)}$ & 0.926$_{(0.016)}$ & 0.873$_{(0.013)}$ \\
ClinTox  & 0.946$_{(0.026)}$ & \textbf{0.935}$_{(0.017)}$ & 0.918$_{(0.004)}$ & 0.921$_{(0.021)}$ & 0.885$_{(0.023)}$ \\
Estrogen & 0.906$_{(0.034)}$ & 0.917$_{(0.050)}$ & \textbf{0.920}$_{(0.046)}$ & 0.912$_{(0.056)}$ & 0.904$_{(0.069)}$ \\
Metstab  & 0.889$_{(0.057)}$ & \textbf{0.891}$_{(0.059)}$ & 0.888$_{(0.063)}$ & 0.875$_{(0.047)}$ & 0.869$_{(0.072)}$ \\
SIDER    & 0.649$_{(0.011)}$ & \textbf{0.655}$_{(0.016)}$ & 0.647$_{(0.023)}$ & 0.646$_{(0.017)}$ & 0.654$_{(0.015)}$ \\
ToxCast  & 0.745$_{(0.003)}$ & \textbf{0.724}$_{(0.008)}$ & 0.723$_{(0.019)}$ & 0.717$_{(0.016)}$ & 0.706$_{(0.015)}$ \\
Tox21    & 0.848$_{(0.017)}$ & \textbf{0.827}$_{(0.028)}$ & 0.823$_{(0.032)}$ & 0.821$_{(0.024)}$ & 0.815$_{(0.016)}$ \\
\midrule
\textit{AVG (Cls)} & 0.843 & \textbf{0.845} & 0.838 & 0.833 & 0.818 \\
\midrule
\multicolumn{6}{c}{\textit{Regression (RMSE $\downarrow$)}} \\
ESOL     & 0.804$_{(0.102)}$ & 0.761$_{(0.046)}$ & \textbf{0.740}$_{(0.054)}$ & 0.803$_{(0.045)}$ & 0.956$_{(0.045)}$ \\
FreeSolv & 2.121$_{(1.025)}$ & \textbf{2.110}$_{(0.959)}$ & 2.192$_{(0.832)}$ & 2.188$_{(0.820)}$ & 2.328$_{(0.919)}$ \\
Lipo     & 0.600$_{(0.012)}$ & \textbf{0.645}$_{(0.023)}$ & 0.652$_{(0.022)}$ & 0.653$_{(0.027)}$ & 0.706$_{(0.014)}$ \\
\midrule
\textit{AVG (Reg)} & 1.175 & \textbf{1.172} & 1.195 & 1.215 & 1.329 \\
\bottomrule
\end{tabular}
\end{sc}
\end{small}
\end{center}
\end{table}
% ==========================================
% Table 2: MoleculeACE Full Ablation
% Unified Font: small, Unified Order: Full -> w/o FP -> 1K -> 67K
% ==========================================
\begin{table}[h]
\caption{Full Ablation Results on MoleculeACE (RMSE $\downarrow$). The column order is consistent with Table~\ref{tab:app_ablation_molnet}. The \textbf{Full} model consistently yields the best stability (lowest RMSE).}
\label{tab:app_ablation_ace}
\begin{center}
\begin{small} % Changed from tiny to small to match Table 1
\begin{sc}
\setlength{\tabcolsep}{2.5pt} % Tighter padding to fit 30 rows in width
\begin{tabular}{lccccc}
\toprule
Task & KPGT & \textbf{CamS (Full)} & w/o FP & 1K Only & 67K Only \\
\midrule
CHEMBL1862 & 0.633 & \textbf{0.600} & 0.635 & 0.624 & 0.609 \\
CHEMBL1871 & 0.605 & \textbf{0.604} & \textbf{0.604} & 0.606 & 0.623 \\
CHEMBL2034 & 0.679 & \textbf{0.619} & 0.679 & \textbf{0.615} & 0.655 \\
CHEMBL204  & \textbf{0.666} & \textbf{0.709} & 0.749 & 0.730 & 0.726 \\
CHEMBL2047 & 0.578 & \textbf{0.519} & 0.578 & \textbf{0.518} & 0.563 \\
CHEMBL214  & 0.652 & \textbf{0.635} & 0.662 & 0.660 & 0.663 \\
CHEMBL2147 & 0.587 & \textbf{0.577} & 0.632 & 0.605 & 0.596 \\
CHEMBL218  & \textbf{0.625} & \textbf{0.632} & 0.687 & 0.655 & 0.646 \\
CHEMBL219  & \textbf{0.718} & \textbf{0.729} & 0.736 & \textbf{0.723} & 0.738 \\
CHEMBL228  & \textbf{0.669} & \textbf{0.669} & 0.679 & 0.676 & 0.713 \\
CHEMBL231  & \textbf{0.610} & \textbf{0.630} & 0.642 & 0.642 & 0.638 \\
CHEMBL233  & \textbf{0.691} & \textbf{0.692} & 0.719 & 0.724 & 0.712 \\
CHEMBL234  & \textbf{0.606} & \textbf{0.624} & 0.630 & 0.643 & 0.645 \\
CHEMBL235  & 0.624 & \textbf{0.612} & 0.629 & \textbf{0.604 }& 0.629 \\
CHEMBL236  & \textbf{0.655} & \textbf{0.669} & 0.709 & 0.708 & 0.704 \\
CHEMBL237$_{E}$ & 0.716 & \textbf{0.684} & 0.695 & 0.712 & 0.737 \\
CHEMBL237$_{K}$ & 0.660 & \textbf{0.659} & \textbf{0.659} & 0.721 & 0.712 \\
CHEMBL238  & \textbf{0.537} & \textbf{0.537} & 0.585 & 0.564 & 0.572 \\
CHEMBL239  & \textbf{0.644} & \textbf{0.647} & 0.672 & 0.676 & 0.655 \\
CHEMBL244  & 0.698 & \textbf{0.696} & 0.726 & 0.723 & 0.725 \\
CHEMBL262  & \textbf{0.627} & \textbf{0.629} & 0.662 & 0.637 & 0.662 \\
CHEMBL264  & 0.574 & \textbf{0.562} & 0.570 & 0.583 & 0.583 \\
CHEMBL2835 & \textbf{0.373} & \textbf{0.384} & 0.392 & 0.385 & \textbf{0.384} \\
CHEMBL287  & 0.706 & \textbf{0.685} & \textbf{0.685} & \textbf{0.683} & 0.742 \\
CHEMBL2971 & \textbf{0.571} & \textbf{0.574} & 0.596 & 0.584 & 0.584 \\
CHEMBL3979 & 0.669 & \textbf{0.639} & 0.672 & 0.652 & 0.655 \\
CHEMBL4005 & 0.559 & \textbf{0.543} & 0.581 & 0.553 & 0.561 \\
CHEMBL4203 & 0.820 & \textbf{0.787} & 0.820 & 0.800 & 0.811 \\
CHEMBL4616 & 0.587 & \textbf{0.538} & 0.565 & 0.553 & 0.565 \\
CHEMBL4792 & \textbf{0.619} & \textbf{0.651} & 0.659 & 0.668 & 0.658 \\
\midrule
\textbf{AVG} & 0.632 & \textbf{0.624} & 0.650 & 0.641 & 0.649 \\
\bottomrule
\end{tabular}
\end{sc}
\end{small}
\end{center}
\end{table}
\end{document}